\documentclass[]{opendatalab}
\usepackage{xcolor}
\usepackage{longtable}
\usepackage{listings}
\usepackage[numbers]{natbib}
\definecolor{codegreen}{rgb}{0,0.6,0}
\definecolor{codegray}{rgb}{0.5,0.5,0.5}
\definecolor{codepurple}{rgb}{0.58,0,0.82}
\definecolor{backcolour}{rgb}{0.95,0.95,0.92}
\definecolor{promptcolor}{HTML}{D1D0F2}
\definecolor{promptcolorheader}{HTML}{bdbcec}
\newcommand{\promptbox}[2]{
\begin{tcolorbox}[
top=0.3em,bottom=0.3em,left=0.5em,right=0.5em,
toptitle=0.3em,bottomtitle=0.2em,boxsep=0pt,
colframe=promptcolorheader,colback=promptcolor!50,boxrule=0.5pt,
]
\footnotesize
\end{tcolorbox}
}
\lstdefinestyle{mystyle}{
    backgroundcolor=\color{backcolour},   
    commentstyle=\color{codegreen},
    keywordstyle=\color{magenta},
    numberstyle=\tiny\color{codegray},
    stringstyle=\color{codepurple},
    basicstyle=\ttfamily\footnotesize,
    breakatwhitespace=false,         
    breaklines=true,                 
    captionpos=b,                    
    keepspaces=true,                 
    numbers=left,                    
    numbersep=5pt,                  
    showspaces=false,                
    showstringspaces=false,
    showtabs=false,                  
    tabsize=2
}

\usepackage{algorithm}
\usepackage{algorithmic}
\usepackage{multirow}
\usepackage{booktabs}
\usepackage{enumitem}
\usepackage{tabularx}
\usepackage{amsmath}

\usepackage{pifont}        

\usepackage{newfloat}
\usepackage{listings}

\title{GTR-VL: Graph Traversal as Visual Chain of Thought for Molecular Structure Recognition}

\author[2\dag]{Jingchao Wang}
\author[1\dag]{Yifan He}
\author[1\dag]{Haote Yang}
\author[3\dag]{Jiang Wu}
\author[4]{Lingli Ge}
\author[1]{Xingjian Wei}
\author[1]{Yinfan Wang}
\author[5]{Linye Li}
\author[6]{Huijie Ao}
\author[7]{Chengjin Liu}
\author[1]{Bin Wang}
\author[1]{Lijun Wu}
\author[1]{Conghui He}

\affiliation[1]{Shanghai Artificial Intelligence Laboratory}
\affiliation[2]{East China Normal University}
\affiliation[3]{Peking University}
\affiliation[4]{Shanghai Jiaotong University}
\affiliation[5]{Tongji University}
\affiliation[6]{Fudan University}
\affiliation[7]{Northwestern Polytechnical University}

\abstract{
Optical Chemical Structure Recognition (OCSR) is essential for converting molecular images into machine-readable formats. While recent vision-language models (VLMs) have shown promise, their image-captioning approach often struggles with complex molecular structures and inconsistent annotations. 
To address these issues, we introduce GTR-VL, featuring two key innovations: (1) the \textit{Graph Traversal as Visual Chain of Thought} mechanism that emulates human reasoning by incrementally parsing molecular graphs through sequential atom-bond predictions, and (2) the data-centric \textit{Faithfully Recognize What You've Seen} principle, which aligns abbreviated structures in images with their expanded annotations.
For hand-drawn OCSR tasks, where datasets lack graph annotations and only provide final SMILES, we apply reinforcement learning using the GRPO method, introducing reward mechanisms like format reward, graph reward, and SMILES reward. This approach significantly enhances performance in hand-drawn recognition tasks through weak supervision.
We developed GTR-1.3M, a large-scale instruction-tuning dataset with corrected annotations, and MolRec-Bench, the first benchmark for fine-grained evaluation of graph-parsing accuracy in OCSR. Our two-stage training scheme involves SFT training for printed images and the GRPO method for transferring capabilities to hand-drawn tasks.
Experiments show that GTR-VL outperforms specialist models, chemistry-domain VLMs, and commercial VLMs on both printed and hand-drawn datasets.
}

\date{\today}
\metadata[Equal contribution]{Jingchao Wang, Yifan He, Haote Yang, Jiang Wu}
\metadata[Project leader]{Jiang Wu}
\correspondence{Conghui He, \email{heconghui@pjlab.org.cn}}

\begin{document}

\maketitle
\section{Introduction}

\begin{figure*}[t]
    \centering
    \includegraphics[width=1.0\linewidth]{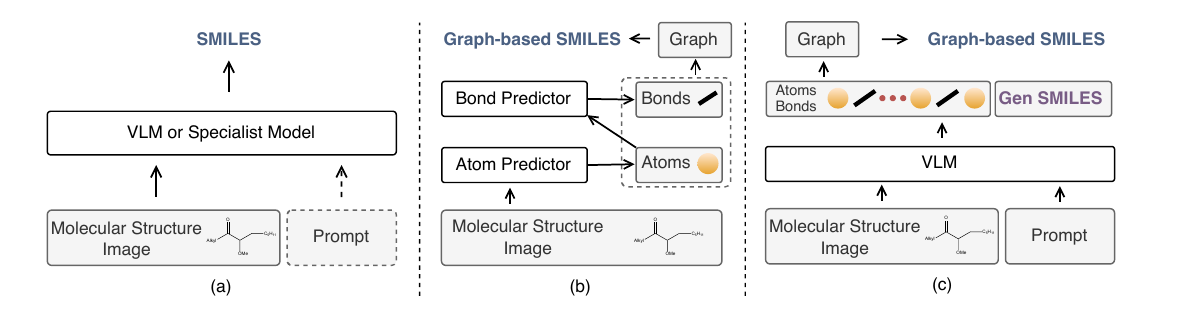}
    \caption{The comparison of three paradigms.
(a) Image-captioning approach: Directly generates SMILES from molecular structure images using either VLMs or specialist models.
(b) Graph-parsing approach: 
Predict atoms and bonds in separate stages to construct a molecular graph, which is then converted to SMILES.
(c) Ours: Jointly generates atoms and bonds to form a molecular graph, followed by SMILES generation. The graph is then used to construct a graph-based SMILES.}
    \label{fig:model_comparison}
\end{figure*}

Modern chemistry's vast knowledge is often stored in chemical molecular formulas, typically as 2D images in papers and patents, limiting machine accessibility (\cite{rajan2020review}). With the rise of Large Language Models (LLMs), converting these images for model training is crucial. Optical Chemical Structure Recognition (OCSR) technology addresses this by converting images into machine-readable formats like SMILES, crucial for digitizing chemical data and advancing AI in chemistry. Early rule-based methods (\cite{filippov2009optical,peryea2019molvec}) were limited to simple cases, and recent deep learning advances have enhanced OCSR capabilities (\cite{rajan2021decimer,fang2024molparser,morin2023molgrapher,qian2023molscribe}). Nonetheless, further improvements are needed for handling large molecules, complex Markush structures (\cite{morin2025markushgrapher}), and hand-drawn formats.

Recently, large Vision-Language Models (VLMs) (\cite{bai2023qwenvlversatilevisionlanguagemodel, chen2024internvl}) have achieved breakthroughs in visual perception (\cite{Liu_2024}), visual question answering (\cite{yu2024mmvetevaluatinglargemultimodal}), and multimodal reasoning (\cite{yue2024mmmuprorobustmultidisciplinemultimodal, gao2025pm4benchparallelmultilingualmultimodal}). They have been applied in fields like medicine (\cite{li2023llavamedtraininglargelanguageandvision}), autonomous driving (\cite{duan2024cityllavaefficientfinetuningvlms}), remote sensing (\cite{pang2024vhmversatilehonestvision, muhtar2024lhrsbotempoweringremotesensing}), and OCR (\cite{wei2024generalocrtheoryocr20}). Recent works such as ChemVLM (\cite{li2025chemvlmexploringpowermultimodal}), ChemDFM-X (\cite{Zhao_2024}), and OCSU \cite{fan2025ocsuopticalchemicalstructure} have applied VLMs to OCSR, treating it as image captioning to generate SMILES strings. However, this approach is less effective than graph-parsing methods, and their performance on OCSR tasks needs improvement, as shown in Table \ref{tab:performance_pr}.

After evaluating existing models, we propose two insights and design principles:
\textbf{(1) Graph Traversal as Visual Chain of Thought:} The Chain of Thought (CoT) technique improves problem-solving by generating intermediate steps. For OCSR tasks, a visual CoT mechanism that recognizes complex molecules step-by-step can enhance performance. Unlike current methods (\cite{morin2023molgrapher,qian2023molscribe,chen2024molnextr}) that predict atoms and bonds separately, our \textit{Graph Traversal as Visual CoT} method interleaves atom and bond predictions in one pass, improving accuracy and consistency.
\textbf{(2) Faithfully Recognize What You've Seen:} Abbreviations like "Ph" for phenyl in molecular images challenge OCSR tasks. Existing methods (\cite{qian2023molscribe,chen2024molnextr}) struggle due to mismatches between images and annotations. We developed a data correction pipeline that aligns annotations with images by representing abbreviations as superatoms, improving model accuracy.
Extensive experiments demonstrate that our principles significantly enhance VLM accuracy and ensure alignment with molecular structure diagrams, which improves interpretability and facilitates manual inspection and editing.

We extended these principles to hand-drawn molecular recognition tasks. Existing datasets lack fine-grained supervision for atoms and bonds, limiting model performance. We addressed this using Reinforcement Learning (RL) with the Group Relative Policy Optimization (GRPO)(\cite{shao2024deepseekmathpushinglimitsmathematical}) method, designing reward mechanisms including format, graph, and SMILES rewards.

Building on these insights, we developed \textbf{GTR-1.3M}, an SFT dataset for VLM OCSR tasks, containing 1.3 million samples with molecular images, a visual CoT process, and final SMILES strings (Figure ~\ref{fig:train}). 
Leveraging these principles, the RL method, and the dataset, we propose a two-stage training scheme and develop \textbf{GTR-VL}, a specialized multimodal model. This model excels in handling complex molecular images and hand-drawn recognition tasks, advancing OCSR technology to better meet practical needs.


The contributions of this paper are as follows:

1. We apply VLM technology to OCSR, introducing two design principles: \textit{Graph Traversal as Visual Chain of Thought} and \textit{Faithfully Recognize What You've Seen}. These principles enhance VLM accuracy and ensure alignment between molecular diagrams and images, improving interpretability and facilitating manual editing.

2. We utilize RL with the GRPO method for molecular structure recognition, introducing reward mechanisms like format, graph, and SMILES rewards. This approach effectively improves performance in recognizing hand-drawn molecular structures with only SMILES annotations.

3. Using these principles, we developed the VLM SFT dataset \textbf{GTR-1.3M} and introduced \textbf{MolRec-Bench}, a benchmark for assessing graph parsing accuracy in OCSR tasks. \textbf{MolRec-Bench} addresses the limitations of SMILES-based evaluations by accurately assessing structures with custom functional groups or abbreviations.

4. Based on the principles, RL method, and dataset, we propose the two-stage training scheme and develop the \textbf{GTR-VL} model, which excels with complex molecular images and hand-drawn recognition tasks.

\section{Preliminary}

\begin{figure}[t]
    \centering
    \includegraphics[width=1.0\linewidth]{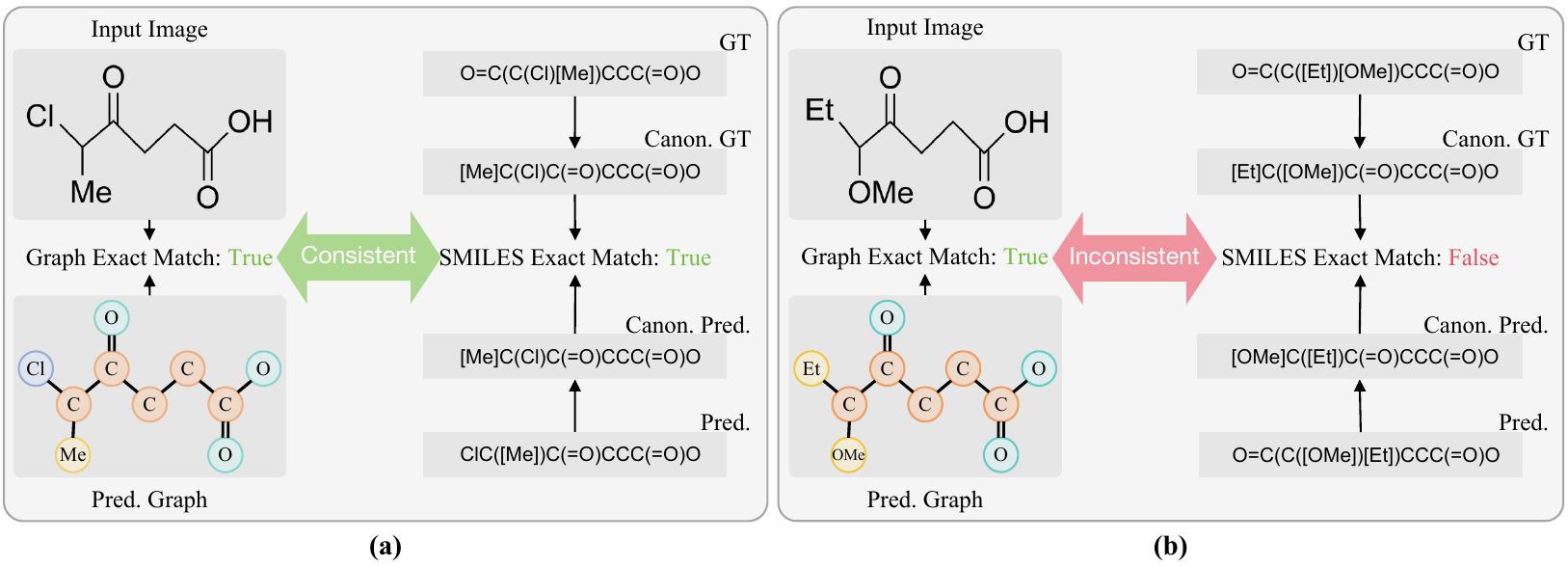}
    \caption{Illustration of limitations in SMILES-based evaluation. (a) The positive example: both the predicted graph and SMILES match with the ground truth. (b) The counterexample: the graph-parsing OCSR algorithm correctly interprets the molecular graph, but the SMILES does not match the ground truth and is \textbf{incorrectly judged} as a prediction error.}
    \label{fig:graph_score}
\end{figure}

\subsection{Image-Captioning vs Graph-Parsing}
\label{sec:ocsr_methods}
Image-captioning methods treat OCSR as an image captioning task, outputting SMILES strings directly. In contrast, graph-parsing methods predict atoms, bonds, and molecular information to construct the graph structure. Graph-parsing offers several advantages: (1) \textbf{Interpretability}: It allows for better algorithm optimization and robustness analysis. (2) \textbf{Manual Verification}: Results align with input images, enabling manual checks and semi-automated annotation. (3) \textbf{Expressive Capability}: It can represent complex structures like Markush structures. (4) \textbf{Performance}: It outperforms image-captioning methods with the same training data. Therefore, we chose graph-parsing for this study.

\subsection{Challenges of Hand-drawn OCSR}

Hand-drawn molecular data poses unique challenges compared to printed depictions: the training data is much scarcer, and annotations usually only provide SMILES strings without atomic coordinates, making graph-parsing methods unsuitable. Approaches like DECIMER(\cite{rajan2023decimer}) must rely on generating synthetic data for training, producing over 100 million synthetic samples to achieve competitive recognition within the image-captioning framework.

\subsection{Limitations of SMILES-based Evaluation}
\label{sec:limitations_of_smiles_based_evaluation}

SMILES (\cite{weininger1988smiles})  is a string-based language for representing molecular structures and reactions, encoding atoms and bonds in character sequences (Appendix D). 
OCSR evaluation datasets typically compare predicted and ground truth SMILES using exact matches or similarity measures.
However, due to canonicalization issues, there are situations where the OCSR algorithm may correctly interpret the molecular graph while the SMILES might not match the ground truth and be incorrectly judged as a prediction error (Figure \ref{fig:graph_score}).
Additionally, studies like MolScribe (\cite{qian2023molscribe}) and MolNexTR (\cite{chen2024molnextr}) replace abbreviations with an asterisk (*) in SMILES, ignoring these elements and resulting in incomplete evaluations (see Appendix B for details).

\section{Method}

\begin{figure}[t]
    \centering
    \includegraphics[width=1.0\linewidth]{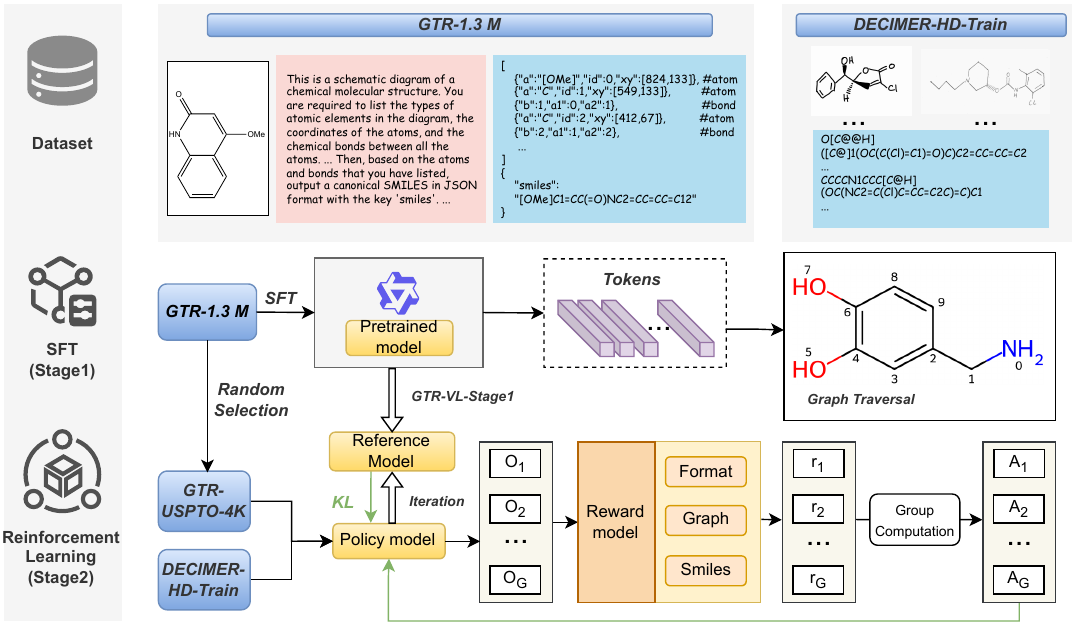}
    \caption{
    Our method is designed to achieve robust recognition across both printed and hand-drawn molecular structures. We begin by performing supervised fine-tuning (SFT) on the large-scale dataset GTR-1.3M to train a base model capable of recognizing printed molecular depictions. To further enhance the model’s ability to handle the challenging domain of hand-drawn inputs, we apply reinforcement learning using the GRPO algorithm on the DECIMER Hand-drawn dataset. This two-stage training pipeline enables the model to generalize effectively across different molecular representation styles.
    }
    \label{fig:train}
\end{figure}

\subsection{Insights and Design Principles}

The graph-parsing OCSR task can be formulated as an image-to-graph generation task. Given an image \(I_{m}\) containing a molecule \(m\), we train a model \(f\) to convert \(I_{m}\) into a molecular structure graph \(G_{m} = \{a_{1}, a_{2}, \ldots, a_{p}, b_{1}, b_{2}, \ldots, b_{q}\}\), where \(a_{i}\) represents the \(i\)-th atom and \(b_{j}\) represents the \(j\)-th bond, i.e., \(G_{m} = f(I_{m})\).
Compared to existing graph-parsing methods, we have two insights and corresponding design principles.


\subsubsection{Graph Traversal as Visual CoT}




Existing graph-parsing methods~(\cite{morin2023molgrapher,qian2023molscribe,chen2024molnextr}) adopt a two-stage approach (Figure~\ref{fig:model_comparison}(b)): first predicting atoms (nodes), then chemical bonds (edges) using classifiers or GNNs. This diverges from human cognition, which naturally alternates attention between atoms and bonds. The two-stage design presents two key limitations:
(1) Atom prediction lacks structural constraints from bonds, increasing ambiguity;
(2) Bond prediction requires global attention over all atoms, leading to higher inference complexity and cost.

We propose \textit{graph traversal as a visual CoT}, a human-inspired method that parses molecular graphs by interleaving atom and bond predictions in a single traversal (Figure~\ref{fig:model_comparison}(c)). As illustrated in Figure~\ref{fig:train}, a depth-first strategy defines the traversal order, alternating atom and bond predictions along the path.
This interleaved process leverages mutual constraints: bonds are predicted based only on previously identified atoms, reducing prediction difficulty. Furthermore, the traversal serves as a visual Chain-of-Thought for VLMs, decomposing complex recognition into structured, sequential sub-tasks, thus enhancing prediction accuracy and consistency.


\subsubsection{Faithfully Recognizing What You've Seen}
\label{sec:faithfully_recognizing}
In molecular structure images from papers and patents, many abbreviated structures (e.g., Ph for phenyl, Pr for propyl, Bu for butyl, as shown in Figure~\ref {fig:prediction_abbr}) are common. While these abbreviations improve readability and conciseness, they present challenges for OCSR tasks.
Existing works (\cite{qian2023molscribe,chen2024molnextr}) use annotations (from MOL files) with fully expanded molecular graphs in their patent training data. This mismatch can confuse models, leading to errors when predicting expanded forms for abbreviations seen in images (Figure ~\ref{fig:prediction_hd}).

We propose treating these abbreviations as "super atoms" rather than expanding them. This \textit{faithfully recognizing what you've seen} approach ensures consistency between images and annotations, optimizing model learning, and significantly enhancing generalization capability.

\subsection{GRPO for hand-drawn OCSR}



As noted earlier, image-captioning methods~(\cite{rajan2020decimer}) face challenges in hand-drawn OCSR due to the need for large-scale annotated data, while graph-parsing methods~(\cite{qian2023molscribe}) are inapplicable because hand-drawn samples lack coordinate information.
However, we observe that although coordinates are missing, topological molecular graphs can still be reconstructed from SMILES. While these coordinate-independent graphs cannot supervise token-level outputs, their structural correctness can be leveraged as a reward signal in GRPO.

Thus, we design a composite reward function integrating response format, molecular graph, and SMILES accuracy. Specifically, we compare the predicted and ground-truth graphs (from SMILES) by computing their maximum common subgraph (MCS). Graph similarity is defined as the MCS size relative to both graphs, and used as the graph-level reward, the formula is as follows where $\textit{N}^{\textit{a}}_{\textit{m}}$, $\textit{N}^{\textit{a}}_{\textit{g}}$, and $\textit{N}^{\textit{a}}_{\textit{p}}$ represent the number of atoms in the MCS, ground truth graph, and prediction graph respectively, and $\textit{N}^{\textit{b}}_{\textit{m}}$, $\textit{N}^{\textit{b}}_{\textit{g}}$, and $\textit{N}^{\textit{b}}_{\textit{p}}$ denote the number of edges respectively.

\begin{equation}
\textit{R}_{\textit{graph}} = \frac{|\textit{N}^{\textit{a}}_{\textit{m}}|}{|\textit{N}^{\textit{a}}_{\textit{g}}| + |\textit{N}^{\textit{a}}_{\textit{p}}|} + \frac{|\textit{N}^{\textit{b}}_{\textit{m}}|}{|\textit{N}^{\textit{b}}_{\textit{g}}| + |\textit{N}^{\textit{b}}_{\textit{p}}|}
\label{eq:reward_graph}
\end{equation}



To enforce output validity, we also include format and SMILES rewards. The final GRPO reward combines these three components as follows.


\begin{equation}
\textit{R}_{\textit{total}} = \textit{R}_{\textit{graph}} + \textit{R}_{\textit{format}} + \textit{R}_{\textit{SMILES}}
\label{eq:total_reward}
\end{equation}

\subsection{Dataset Construction}
\label{sec:dataset_construction}



Building on these insights, we developed \textbf{GTR-1.3M}, a specialized SFT dataset for VLM-based OCSR tasks. Following MolScribe and MolNexTR, \textbf{GTR-1.3M} is composed of two parts:
(1) \textbf{GTR-PubChem-1M}: We selected 1 million molecular SMILES from the PubChem database and used the Indigo tool to convert them into molecular images. 
(2) \textbf{GTR-USPTO-351K}: This subset was created from USPTO-680K. 
We developed a data correction pipeline to correct and filter abbreviated structures in these samples, obtained 351k high-quality samples, and formed the \textbf{GTR-USPTO-351K} subset. Please refer to Appendix A for more details.

\subsection{Two-Stage Training}

Based on the design principles, data selection, and reinforcement learning framework, we adopt a two-stage training strategy (Figure~\ref{fig:train}).

\textbf{Stage~1:} We perform SFT on the \textbf{GTR-1.3M} dataset to teach the VLM a visual CoT mechanism via molecular graph traversal. The model first constructs a molecular graph from the input image, then predicts the SMILES string. During inference, the output is parsed to recover the graph and generate SMILES via rule-based decoding. We denote SMILES directly produced by the VLM as generated SMILES, and those derived from parsed graphs as graph-based SMILES. The resulting model is referred to as \textbf{GTR-VL-stage1}.

\textbf{Stage2:} We apply GRPO using a mixture of printed and hand-drawn data: (1) \textbf{GTR-USPTO-4K}, sampled from \textbf{GTR-USPTO-351K}, and (2) \textbf{DECIMER-HD-Train} (4070 samples following (\cite{oldenhof2024atom})). In Epoch 1, both the policy and reference models are initialized with \textbf{GTR-VL-stage1}; in subsequent epochs, they are updated using the trained model from the previous epoch. The final model is denoted as \textbf{GTR-VL-stage2} or \textbf{GTR-VL}.




\begin{table*}[t]
\centering
\begin{tabularx}{\textwidth}{c*{6}{>{\centering\arraybackslash}X}}
\specialrule{1.5pt}{0pt}{2pt}
\multirow{3}{*}{\textbf{Model}} 
& \multicolumn{3}{c}{\textbf{MolRec-Abb}} 
& \multicolumn{3}{c}{\textbf{MolRec-USPTO}} \\
\cmidrule(r){2-4} \cmidrule(l){5-7}
& \textbf{\texttt{Gen-SMILES}} & \textbf{\texttt{Gra-SMILES}} & \textbf{\texttt{Graph}}
& \textbf{\texttt{Gen-SMILES}} & \textbf{\texttt{Gra-SMILES}} & \textbf{\texttt{Graph}}  \\
\specialrule{0.5pt}{2pt}{2pt}

MolScribe\textsuperscript{†}  & 20.11 & 19.39 & 19.82 & 71.77 & 72.03 & 72.25         \\
MolScribe\textsuperscript{§}  & 72.20 & 69.63 & 70.60 & 85.49 & 84.66 & 86.23         \\
MolNexTR\textsuperscript{†}    & 19.76 & 19.00 & 19.47 & 71.75 & 71.90 & 72.14         \\
MolNexTR\textsuperscript{§}    & 74.60 & 70.98 & 71.85 & 86.54 & 85.30 & 86.96         \\
\specialrule{0.5pt}{1pt}{1pt}

OCSU                        & 0.40         & -             & -             & 1.71          & -             & -             \\
ChemVLM                       & 4.18         & -             & -             & 42.52         & -             & -             \\
ChemDFM-X                    & 0.76         & -             & -             & 26.94         & -             & -             \\
\specialrule{0.5pt}{1pt}{1pt}

GPT-4o\textsuperscript{*}     & 0.60 & - & - & 1.60 & - & - \\
GPT-4o\textsuperscript{‡}     & 0.60 & 0.00 & 0.00 & 0.60 & 0.00 & 0.00 \\
GPT-4o-mini\textsuperscript{*}     & 0.20 & - & - & 0.20 & - & - \\
GPT-4o-mini\textsuperscript{‡}     & 0.00 & 0.00 & 0.00 & 0.00 & 0.00 & 0.00 \\
Qwen-VL-max\textsuperscript{*}  & 0.20 & - & - & 1.40 & - & - \\
Qwen-VL-max\textsuperscript{‡} & 0.20 & 0.00 & 0.00 & 1.00 & 0.00  & 0.00  \\

\specialrule{0.5pt}{1pt}{1pt}

GTR-VL-Stage1 (Ours)           &\textbf{84.50}&\textbf{84.50} &\textbf{85.49} &91.19 &\textbf{91.67} & \textbf{93.45}\\
GTR-VL-Stage2 (Ours) &   81.67 & 82.33 &  82.84 & \textbf{91.31}  & 91.17 &   91.28 \\

\specialrule{1.5pt}{0pt}{2pt}

\end{tabularx}

\caption{
Quantitative results are reported across three exact-match metrics on two sub-benchmarks. 
The highest value for each metric is marked in bold.
§ indicates trained on \textbf{GTR-1.3M}; † indicates officially released checkpoint; ‡ models first predict the graph then SMILES; * models directly predict SMILES.
}
\label{tab:performance_pr}
\end{table*}

\begin{table*}[htbp]
\centering
\label{tab:decimer_chempix}
\begin{tabularx}{\textwidth}{ccccccccc}
\specialrule{1.5pt}{0pt}{2pt}
\multirow{2}{*}{\textbf{Model}} & \multicolumn{4}{c}{\textbf{ DECIMER-HD-Test}} & \multicolumn{4}{c}{\textbf{ChemPix}} \\
\cmidrule(lr){2-5} \cmidrule(lr){6-9}
 & \textbf{\texttt{SMILES}} & \textbf{\texttt{T = 1}} & \textbf{\texttt{T Mean}} & \textbf{\texttt{Graph}} & \textbf{\texttt{SMILES}} & \textbf{\texttt{T = 1}} & \textbf{\texttt{T Mean}} & \textbf{\texttt{Graph}} \\
\midrule
AtomLenz+EditKT$^{\ast}$ (AE)   & 28.00 & 33.69 & 48.40 & 29.76 & 39.80 & 48.29 & 60.50 & 48.29 \\
DECIMER FT (v2.2)    & 56.48 & 61.79 & 72.90 & 61.10 &  55.46 &  58.08 &  75.00 & 57.75 \\
DECIMER FT (v2.2) + AE    & 56.48 & 61.79 & 73.20 & 61.10 & 55.46 & 58.08 & 75.09 & 57.75 \\
DECIMER HD   & 69.94 & 73.28 & 82.92 & 71.71 & 44.37 & 50.24 &  68.16 &  49.76 \\
DECIMER HD + AE  & 69.94 & 73.28 & 83.00 & 71.71 & 44.37 & 50.24 & 68.36 & 49.76 \\

\specialrule{0.5pt}{1pt}{1pt}
GTR-VL-Stage1 (Ours) & 9.53 & 9.92 & 26.83 & 9.53 & 22.02 & 23.16 & 40.73 & 22.02 \\
GTR-VL-Stage2 (Ours) & \textbf{75.44} & \textbf{80.45} & \textbf{86.05} & \textbf{75.44} & \textbf{86.13} & \textbf{86.62} & \textbf{91.28} & \textbf{86.13} \\
\specialrule{1.5pt}{0pt}{2pt}

\end{tabularx}

\caption{Performance comparison on the DECIMER Hand-drawn and ChemPix datasets. AE refers to the method proposed in \cite{oldenhof2024atom}. DECIMER+AE denotes a combined approach that integrates DECIMER with AE.
\textbf{\texttt{SMILES}} indicates the proportion of samples with an exact match of the predicted and ground-truth SMILES strings.  
\textbf{\texttt{T Mean}} denotes the average Tanimoto similarity across all samples.
\textbf{\texttt{T=1}} indicates the proportion of samples with Tanimoto similarity = 1.  
}
\label{tab:performance_hd}
\end{table*}

\subsection{Benchmark and Metrics}

We developed \textbf{MolRec-Bench} to overcome the limitations of existing SMILES-based evaluation datasets as detailed in Section 3.2. This benchmark evaluates not only molecular graph structures but also complex scenarios like Markush structures.
\textbf{MolRec-Bench} comprises two subsets:
(1) \textbf{MolRec-USPTO:} Based on USPTO (\cite{Rajan2020}), it includes 5,423 molecular images from USPTO patents.
(2) \textbf{MolRec-Abb:} Derived from MolGrapher (\cite{morin2023molgrapher}), it features 9,311 molecular images with abbreviated superatoms from USPTO\_10K\_abb.
The construction of \textbf{MolRec-Bench} is followed as the \textbf{GTR-USPTO-351K} (Section~\ref{sec:dataset_construction} and Appendix A). Each sample contains the original molecular image, the corrected molecular graph, and the corrected SMILES.

For \textbf{MolRec-Bench}, we defined three evaluation metrics:
(1) \textbf{\texttt{Gen-SMILES}}: Calculates the exact match ratio by comparing canonicalized ground truth SMILES with predicted SMILES, designed for image-captioning-based OCSR methods.
(2) \textbf{\texttt{Gra-SMILES}}: Similar to \texttt{Gen\_SMILES} but uses SMILES generated from the predicted molecular graph, suited for graph-parsing OCSR methods.
(3) \texttt{Graph}: To compensate for the shortcomings (as mentioned in Section \ref{sec:limitations_of_smiles_based_evaluation}) of the first two measurement methods, we propose a graph-based measurement method, which measures the exact match ratio between the ground truth and predicted graphs for graph-parsing OCSR methods. This method can handle the matching of Markush structures more accurately, thereby enabling more precise evaluation. Details are shown in Appendix B.

\section{Experiments}

\subsection{Experiment Setup}

\subsubsection{Baselines on printed molecule images}
\label{sec:baseline}

To evaluate our method on printed molecule images, we selected three types of baseline models: 
(1)Specialist Models: These are tailored for molecular structure recognition, exemplified by MolScribe (\cite{qian2023molscribe}) and MolNexTR (\cite{chen2024molnextr}). We evaluated both the open-source checkpoints and versions trained from scratch using our \textbf{GTR-1.3M} dataset.
(2) Open-source VLMs in the Chemical Domain: These fine-tuned models, such as ChemVLM (\cite{li2025chemvlmexploringpowermultimodal}), ChemDFM-X (\cite{Zhao_2024}), and OCSU (\cite{fan2025ocsuopticalchemicalstructure}), generate SMILES directly. We tested them on MolRec-Bench, reporting only the \textbf{\texttt{Gen-SMILES}}  using prompts from their papers.
(3) Proprietary General-purpose VLMs: 
We compared models like Qwen-VL-Max-2025-04-08 (\cite{qwen25vl}), GPT-4o-mini-2024-07-18 (\cite{openai_gpt4o-mini}), and GPT-4o-2024-08-06 (\cite{hello_gpt-4o_2024}). Using two prompt sets, we evaluated their ability to generate SMILES directly (\textbf{\texttt{Gen-SMILES}}) and via graph parsing (\textbf{\texttt{Gra-SMILES}}). Due to interface costs, we randomly selected 500 samples from \textbf{MolRec-Abb} and \textbf{MolRec-USPTO} for evaluation.


\begin{figure}[t]
    \centering
    \begin{minipage}[t]{0.46\linewidth}
        \centering
        \includegraphics[width=\linewidth]{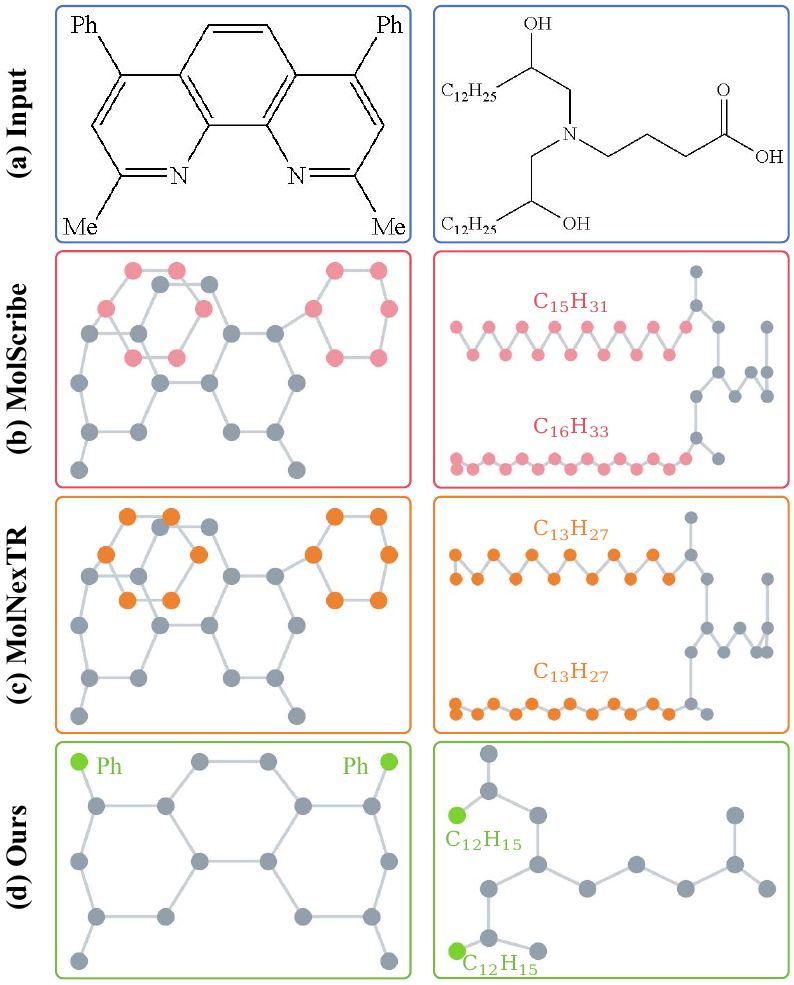}
        \captionof{figure}{Comparison of model predictions on molecular images with abbreviated structures. Our model accurately retains abbreviations as superatoms (d) compared with (b) and (c).}
        \label{fig:prediction_abbr}
    \end{minipage}
    \hfill
    \begin{minipage}[t]{0.52\linewidth}
        \centering
        \includegraphics[width=\linewidth]{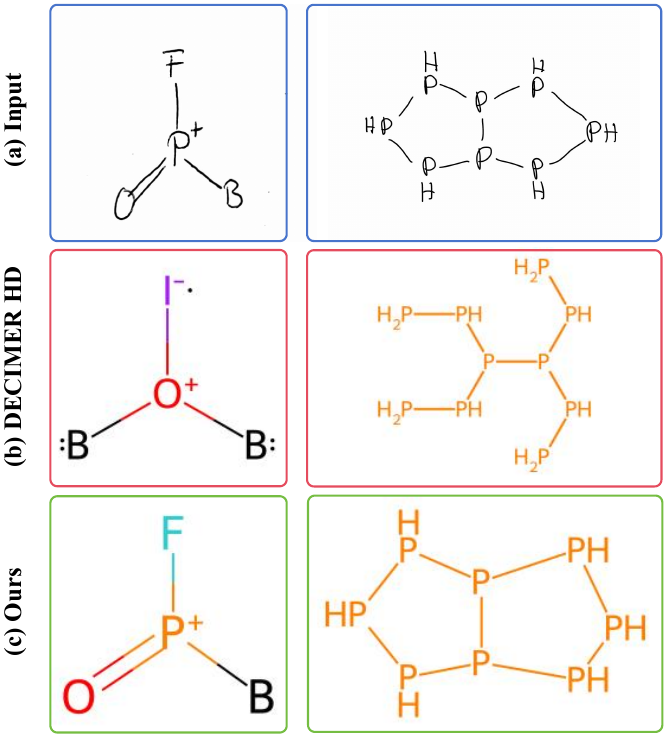}
        \captionof{figure}{Comparison of model predictions on hand-drawn molecular images. Compared with DECIMER (b), our method (c) achieves more accurate predictions.}
        \label{fig:prediction_hd}
    \end{minipage}
\end{figure}

\subsection{Main Results}

\subsubsection{Training Details}

Both the SFT and GRPO (\cite{shao2024deepseekmathpushinglimitsmathematical}) training were performed on 32 NVIDIA A100 GPUs using the AdamW optimizer, with a peak learning rate of \(1.6 \times 10^{-4}\) and \(1 \times 10^{-5}\), respectively. We applied cosine decay for the learning rate and a linear warm-up for the first 10\% of iterations to stabilize training. The batch size per GPU for SFT and GRPO was 2 and 4, respectively. The gradient accumulation steps for SFT and GRPO were 16 and 1, resulting in an effective batch size of 1024 and 128. All model parameters were updated during training. 




\begin{minipage}[t]{0.33\textwidth}
    \centering
    \begin{tabular}{c c}
        \specialrule{1.5pt}{1pt}{1pt}  
        \textbf{CoT} & \textbf{\texttt{Gen\_SMILES}}  \\
        \specialrule{0.5pt}{1pt}{1pt}  
        \ding{55}             & 66.54              \\
        \ding{51}             & 68.85              \\
        \specialrule{1.5pt}{1pt}{1pt}  
    \end{tabular}
    \captionof{table}{The comparison of performances of whether training with visual CoT strategies or not in stage 1.}
    \label{tab:cot}
\end{minipage}
\hfill
\begin{minipage}[t]{0.6\textwidth}
    \centering
    \begin{tabular}{c c c}
        \specialrule{1.5pt}{1pt}{1pt}  
        \textbf{CoT Strategy} & \textbf{\texttt{Gra\_SMILES}} & \textbf{\texttt{Graph}} \\
        \specialrule{0.5pt}{1pt}{1pt}  
        Atoms-then-bonds            & 79.02     & 80.15         \\
        Graph-traversal (ours)      & 81.88     & 83.26         \\
        \specialrule{1.5pt}{1pt}{1pt}  
    \end{tabular}
    \captionof{table}{The performance of our models under different CoT strategies. The Atoms-then-Bonds approach indicates that the model first predicts atoms followed by bonds separately during Stage 1.}
    \label{tab:strategy}
\end{minipage}

\begin{table*}[h]
\centering
\setlength{\tabcolsep}{1mm}
\begin{tabularx}{\textwidth}{*{12}{>{\centering\arraybackslash}X}}
\specialrule{1.5pt}{0pt}{2pt}
\multirow{2}{*}{\textbf{Strategy}} &

\multicolumn{3}{c}{\textbf{Reward / Loss}} &
\multicolumn{4}{c}{\textbf{DECIMER-HD-Test}} &
\multicolumn{4}{c}{\textbf{ChemPix}} \\
\cmidrule(lr){2-4} \cmidrule(lr){5-8} \cmidrule(lr){9-12}
& \textbf{Format} & \textbf{SMILES} & \textbf{Graph}
& \textbf{\texttt{SMILES}} & \textbf{\texttt{T = 1}} & \textbf{\texttt{T Mean}} & \textbf{\texttt{Graph}}
& \textbf{\texttt{SMILES}} & \textbf{\texttt{T = 1}} & \textbf{\texttt{T Mean}} & \textbf{\texttt{Graph}} \\
\specialrule{0.5pt}{2pt}{2pt}

SFT & \ding{51} & \ding{51} &  &
56.09 & 59.04 & 71.68 & 56.19 &
55.14 & 55.79 & 74.14 &  55.14 \\

GRPO  & \ding{51} & \ding{51}  & &
11.00 & 11.49 & 28.72 & 11.00 &
22.84 & 24.31 & 42.35 & 22.84 \\

GRPO  & \ding{51} & \ding{51} & \ding{51} &
\textbf{75.64} & \textbf{79.96} &  \textbf{86.09} & \textbf{75.64} &
\textbf{82.06} &  \textbf{83.52} & \textbf{88.39} &  \textbf{82.06} \\

\specialrule{1.5pt}{2pt}{0pt}
\end{tabularx}

\caption{
Impact of Graph Supervision on the hand-drawn OCSR task.
Results demonstrate that graph supervision substantially improves transfer from printed to hand-drawn domains. While SFT-based transfer (using only SMILES) outperforms GRPO without graph supervision, both are clearly inferior to GRPO with full graph supervision.
}
\label{tab:ablation_study}
\end{table*}

\subsubsection{Overall Performance}

For the printed OCSR task, we compared baseline models from Section~\ref{sec:baseline} with \textbf{GTR-VL}, as shown in Table \ref{tab:performance_pr}.
For \textbf{MolRec-USPTO} and \textbf{MolRec-Abb}, \textbf{GTR-VL-Stage1} achieved the highest performance in all 3 metrics. 
On \textbf{MolRec-Abb}, it surpassed the second-best model by 9 percentage points in all 3 metrics, highlighting its strength in handling molecular images with abbreviations.
Closed-source general-purpose VLMs like GPT-4o (\cite{hello_gpt-4o_2024}) and Qwen (\cite{qwen25vl}), despite excelling in general vision-language tasks, performed poorly on \textbf{MolRec-Bench}, indicating that OCSR hasn't been a focus in their development.
Existing chemical-domain VLMs using image-captioning approaches lack the flexibility of graph-based methods and often underperform on \texttt{SMILES} metrics compared to earlier specialist models. This underscores the effectiveness of the approach we present. 
Although the performance declined after the second stage of GRPO training (which expanded the model's recognition capabilities to hand-drawn data), it still maintained a lead over existing methods. All of our used Prompts are listed in Appendix E.

For the hand-drawn OCSR task, as shown in Table~\ref{tab:performance_hd}, \textbf{GTR-VL-Stage1} (trained with SFT only) performs significantly worse than baseline models on both datasets, with \texttt{SMILES} reaching only 8.84\% and \texttt{Graph} only 9.53\% on \textbf{DECIMER-HD-Test}, indicating that the model trained only with \textbf{GTR-1.3M} cannot be directly applied to the task of handwritten chemical formula recognition.
However, after introducing the proposed reward function and applying GRPO optimization on top of SFT, \textbf{GTR-VL} achieves a dramatic performance boost across all metrics. For example, on \textbf{DECIMER-HD-Test}, \texttt{SMILES} jumps from 8.84\% to 75.34\%, and Graph from 9.53\% to 75.44\%; similarly, on ChemPix, \texttt{SMILES} and \texttt{Graph} score from 22.02\% to 86.13\%, respectively. This fully demonstrates the effectiveness of the reward function we proposed.

\subsubsection{Further Analysis}

 As shown in Table~\ref{tab:performance_pr}, MolScribe and MolNexTR both perform well on \textbf{MolRec-USPTO}, achieving over 70\% accuracy in \texttt{Gen\_SMILES} and \textbf{\texttt{Gra-SMILES}}. However, their performance drops below 20\% on \textbf{MolRec-Abb}, primarily due to mismatches between abbreviated inputs and expanded ground truth graphs (see Section~\ref{sec:faithfully_recognizing}).

To ensure fairness, we retrained both models from scratch using the \textbf{GTR-1.3M} dataset and official code. Their performance improved notably on \textbf{MolRec-Bench}, validating the quality of our data and pipeline. Nevertheless, \textbf{GTR-VL}, powered by QwenVL~(\cite{qwen25vl}), consistently outperforms them. The larger performance gap on \textbf{MolRec-Abb} is attributed to its greater use of abbreviations, making it more challenging.

To more intuitively demonstrate the superiority of our approach, as shown in Figure \ref{fig:prediction_abbr} and \ref{fig:prediction_hd}, we provide a visual comparison of the prediction results of \textbf{MolScribe}, \textbf{MolNexTR}, and \textbf{GTR-VL-Stage1} on the \textbf{MolRec-Abb} dataset, as well as a visual comparison between \textbf{DECIMER HD} and \textbf{GTR-VL-Stage2} on the \textbf{DECIMER-HD-Test} dataset. As can be observed from the figure, our method exhibits clear advantages in both abbreviation recognition and hand-drawn molecular structure prediction.

Compared to their originally reported results, both models show significant degradation. This stems from their evaluation protocol, which treats non-expandable Markush groups as wildcard “*”, potentially inflating accuracy. Additional results and analysis are provided in Appendix~C.

\begin{figure}
    \centering
    \includegraphics[width=0.7\linewidth]{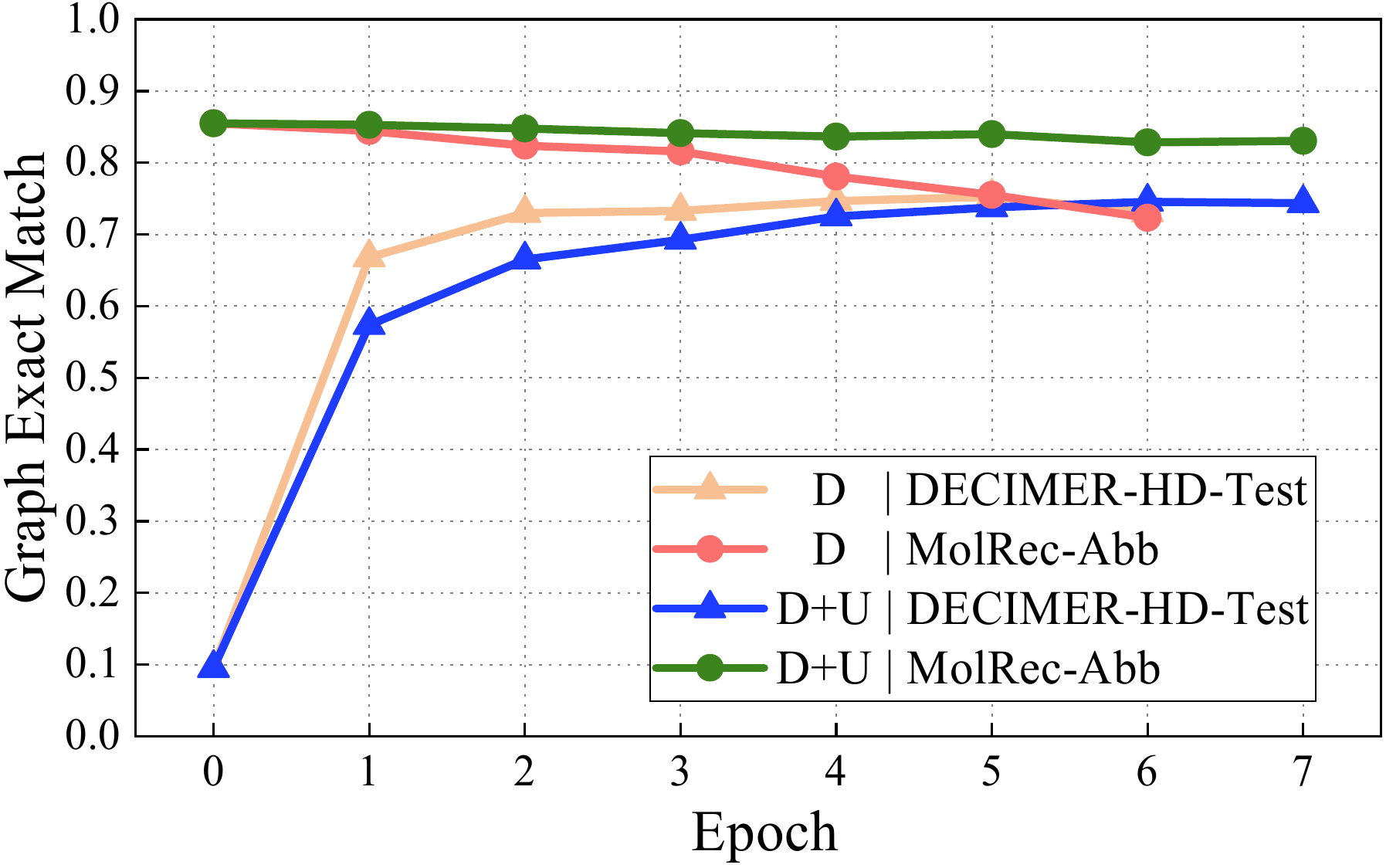}
    \caption{Evolution of Graph Exact Match performance over training epochs for models trained on D or D+U. D: DECIMER-HD-Train only; D+U: joint training with GTR-USPTO-4K.}
    \label{fig:grpo_epoch}
\end{figure}

\subsubsection{Ablation Study}

To further validate the effectiveness of the key design components in our proposed method, we conducted the following ablation experiments:

(1) \textbf{Effectiveness of CoT}: We evaluated the impact of the CoT paradigm using the \textbf{GTR-USPTO-351K} subset. Models were trained with either our \textit{graph traversal as visual CoT} strategy or a baseline that directly predicts SMILES without CoT. As shown in Table~\ref{tab:cot}, the CoT strategy yields a 2.31\% improvement on \texttt{Gen\_SMILES}, indicating that decomposing the recognition task via CoT enhances both performance and stability.

(2) \textbf{Choice of CoT Strategies}: We compared our \textit{graph traversal as visual CoT} with an \textit{Atom-then-bonds} approach, which sequentially predicts atom nodes, bond edges, and methods in MolScribe~(\cite{qian2023molscribe}) and MolNexTR~(\cite{chen2024molnextr}). As shown in Table~\ref{tab:strategy}, our approach achieves gains of 2.86\% and 3.11\% on \textbf{\texttt{Gra-SMILES}}, validating the effectiveness of the graph traversal strategy.

(3) \textbf{GRPO Reward Scheme}: To assess our reward scheme, we compared: (a) SFT with SMILES supervision only, (b) GRPO with rewards from response format and SMILES, and (c) GRPO with rewards from response format, SMILES, and graph. As shown in Table~\ref{tab:ablation_study}, (b) struggles to generalize from \textbf{GTR-1.3M} to \textbf{DECIMER-HD-Train} due to the absence of graph-level guidance. While (a) performs reasonably well, (c) significantly outperforms both, highlighting the importance of incorporating graph-based rewards.

(4) \textbf{GRPO Training Domain}: We evaluated domain impact by comparing (a) GRPO training on hand-drawn data only and (b) joint training on hand-drawn and printed data. Model (a) performs well on \textbf{DECIMER-HD-Test} but poorly on \textbf{MolRec-Abb}. In contrast, (b) improves generalization to \textbf{DECIMER-HD-Test} while retaining moderate performance on \textbf{MolRec-Abb}. By Epoch 6, (b) matches (a)'s performance on \textbf{DECIMER-HD-Test} (Figure \ref{fig:grpo_epoch}).

\section{Conclusion}
This paper introduces GTR-VL, a visual large language model designed for OCSR tasks, along with its accompanying SFT training dataset, GTR-1.3M. GTR-VL is developed based on two core principles: \textit{Graph Traversal as Visual Chain-of-Thought} and \textit{Faithfully Recognizing What You've Seen}. The experiments demonstrate that these innovative concepts not only significantly enhance the applicability and flexibility of OCSR models but also effectively improve model performance. Furthermore, GTR-VL robustly handles challenging scenarios, such as Markush structures. Additionally, our release of the MolRec-Bench addresses the gap in existing OCSR evaluation sets by providing a means to assess molecular graph structure parsing results. We anticipate that GTR-VL and the related dataset will drive OCSR technology toward meeting real-world needs more effectively, thereby advancing the fields of cheminformatics and AI for Science.


\clearpage
\newpage
\bibliographystyle{plainnat}
\setcitestyle{numbers}
\bibliography{paper}

@misc{hello_gpt-4o_2024,
  title = {Hello GPT-4o},
  author = {OpenAI},
  howpublished = {\url{https://openai.com/index/hello-gpt-4o/}},
  year={2024}
}

@article{rajan2024advancements,
  title={Advancements in hand-drawn chemical structure recognition through an enhanced DECIMER architecture},
  author={Rajan, Kohulan and Brinkhaus, Henning Otto and Zielesny, Achim and Steinbeck, Christoph},
  journal={Journal of Cheminformatics},
  volume={16},
  number={1},
  pages={78},
  year={2024},
  publisher={Springer}
}

@article{rajan2023decimer,
  title={DECIMER. ai: an open platform for automated optical chemical structure identification, segmentation and recognition in scientific publications},
  author={Rajan, Kohulan and Brinkhaus, Henning Otto and Agea, M Isabel and Zielesny, Achim and Steinbeck, Christoph},
  journal={Nature communications},
  volume={14},
  number={1},
  pages={5045},
  year={2023},
  publisher={Nature Publishing Group UK London}
}

@article{rajan2020decimer,
  title={DECIMER: towards deep learning for chemical image recognition},
  author={Rajan, Kohulan and Zielesny, Achim and Steinbeck, Christoph},
  journal={Journal of Cheminformatics},
  volume={12},
  number={1},
  pages={65},
  year={2020},
  publisher={Springer}
}

@article{oldenhof2024atom,
  title={Atom-Level Optical Chemical Structure Recognition with Limited Supervision},
  author={Oldenhof, Martijn and De Brouwer, Edward and Arany, Adam and Moreau, Yves},
  journal={arXiv preprint arXiv:2404.01743},
  year={2024}
}

@misc{bai2023qwenvlversatilevisionlanguagemodel,
      title={Qwen-VL: A Versatile Vision-Language Model for Understanding, Localization, Text Reading, and Beyond}, 
      author={Jinze Bai and Shuai Bai and Shusheng Yang and Shijie Wang and Sinan Tan and Peng Wang and Junyang Lin and Chang Zhou and Jingren Zhou},
      year={2023},
      eprint={2308.12966},
      archivePrefix={arXiv},
      primaryClass={cs.CV},
      url={https://arxiv.org/abs/2308.12966}, 
}

@article{chen2024molnextr,
  title={MolNexTR: a generalized deep learning model for molecular image recognition},
  author={Chen, Yufan and Leung, Ching Ting and Huang, Yong and Sun, Jianwei and Chen, Hao and Gao, Hanyu},
  journal={Journal of Cheminformatics},
  volume={16},
  number={1},
  pages={141},
  year={2024},
  publisher={Springer}
}

@article{qian2023molscribe,
  title={MolScribe: robust molecular structure recognition with image-to-graph generation},
  author={Qian, Yujie and Guo, Jiang and Tu, Zhengkai and Li, Zhening and Coley, Connor W and Barzilay, Regina},
  journal={Journal of chemical information and modeling},
  volume={63},
  number={7},
  pages={1925--1934},
  year={2023},
  publisher={ACS Publications}
}

@inproceedings{chen2024internvl,
  title={Internvl: Scaling up vision foundation models and aligning for generic visual-linguistic tasks},
  author={Chen, Zhe and Wu, Jiannan and Wang, Wenhai and Su, Weijie and Chen, Guo and Xing, Sen and Zhong, Muyan and Zhang, Qinglong and Zhu, Xizhou and Lu, Lewei and others},
  booktitle={Proceedings of the IEEE/CVF conference on computer vision and pattern recognition},
  pages={24185--24198},
  year={2024}
}

@misc{yue2024mmmuprorobustmultidisciplinemultimodal,
      title={MMMU-Pro: A More Robust Multi-discipline Multimodal Understanding Benchmark}, 
      author={Xiang Yue and Tianyu Zheng and Yuansheng Ni and Yubo Wang and Kai Zhang and Shengbang Tong and Yuxuan Sun and Botao Yu and Ge Zhang and Huan Sun and Yu Su and Wenhu Chen and Graham Neubig},
      year={2024},
      eprint={2409.02813},
      archivePrefix={arXiv},
      primaryClass={cs.CL},
      url={https://arxiv.org/abs/2409.02813}, 
}

@misc{gao2025pm4benchparallelmultilingualmultimodal,
      title={PM4Bench: A Parallel Multilingual Multi-Modal Multi-task Benchmark for Large Vision Language Model}, 
      author={Junyuan Gao and Jiahe Song and Jiang Wu and Runchuan Zhu and Guanlin Shen and Shasha Wang and Xingjian Wei and Haote Yang and Songyang Zhang and Weijia Li and Bin Wang and Dahua Lin and Lijun Wu and Conghui He},
      year={2025},
      eprint={2503.18484},
      archivePrefix={arXiv},
      primaryClass={cs.CV},
      url={https://arxiv.org/abs/2503.18484}, 
}

@misc{li2023llavamedtraininglargelanguageandvision,
      title={LLaVA-Med: Training a Large Language-and-Vision Assistant for Biomedicine in One Day}, 
      author={Chunyuan Li and Cliff Wong and Sheng Zhang and Naoto Usuyama and Haotian Liu and Jianwei Yang and Tristan Naumann and Hoifung Poon and Jianfeng Gao},
      year={2023},
      eprint={2306.00890},
      archivePrefix={arXiv},
      primaryClass={cs.CV},
      url={https://arxiv.org/abs/2306.00890}, 
}

@misc{duan2024cityllavaefficientfinetuningvlms,
      title={CityLLaVA: Efficient Fine-Tuning for VLMs in City Scenario}, 
      author={Zhizhao Duan and Hao Cheng and Duo Xu and Xi Wu and Xiangxie Zhang and Xi Ye and Zhen Xie},
      year={2024},
      eprint={2405.03194},
      archivePrefix={arXiv},
      primaryClass={cs.CV},
      url={https://arxiv.org/abs/2405.03194}, 
}

@misc{pang2024vhmversatilehonestvision,
      title={VHM: Versatile and Honest Vision Language Model for Remote Sensing Image Analysis}, 
      author={Chao Pang and Xingxing Weng and Jiang Wu and Jiayu Li and Yi Liu and Jiaxing Sun and Weijia Li and Shuai Wang and Litong Feng and Gui-Song Xia and Conghui He},
      year={2024},
      eprint={2403.20213},
      archivePrefix={arXiv},
      primaryClass={cs.CV},
      url={https://arxiv.org/abs/2403.20213}, 
}

@misc{muhtar2024lhrsbotempoweringremotesensing,
      title={LHRS-Bot: Empowering Remote Sensing with VGI-Enhanced Large Multimodal Language Model}, 
      author={Dilxat Muhtar and Zhenshi Li and Feng Gu and Xueliang Zhang and Pengfeng Xiao},
      year={2024},
      eprint={2402.02544},
      archivePrefix={arXiv},
      primaryClass={cs.CV},
      url={https://arxiv.org/abs/2402.02544}, 
}

@misc{wei2024generalocrtheoryocr20,
      title={General OCR Theory: Towards OCR-2.0 via a Unified End-to-end Model}, 
      author={Haoran Wei and Chenglong Liu and Jinyue Chen and Jia Wang and Lingyu Kong and Yanming Xu and Zheng Ge and Liang Zhao and Jianjian Sun and Yuang Peng and Chunrui Han and Xiangyu Zhang},
      year={2024},
      eprint={2409.01704},
      archivePrefix={arXiv},
      primaryClass={cs.CV},
      url={https://arxiv.org/abs/2409.01704}, 
}

@article{Liu_2024,
   title={OCRBench: on the hidden mystery of OCR in large multimodal models},
   volume={67},
   ISSN={1869-1919},
   url={http://dx.doi.org/10.1007/s11432-024-4235-6},
   DOI={10.1007/s11432-024-4235-6},
   number={12},
   journal={Science China Information Sciences},
   publisher={Springer Science and Business Media LLC},
   author={Liu, Yuliang and Li, Zhang and Huang, Mingxin and Yang, Biao and Yu, Wenwen and Li, Chunyuan and Yin, Xu-Cheng and Liu, Cheng-Lin and Jin, Lianwen and Bai, Xiang},
   year={2024},
   month=dec }

@misc{yu2024mmvetevaluatinglargemultimodal,
      title={MM-Vet: Evaluating Large Multimodal Models for Integrated Capabilities}, 
      author={Weihao Yu and Zhengyuan Yang and Linjie Li and Jianfeng Wang and Kevin Lin and Zicheng Liu and Xinchao Wang and Lijuan Wang},
      year={2024},
      eprint={2308.02490},
      archivePrefix={arXiv},
      primaryClass={cs.AI},
      url={https://arxiv.org/abs/2308.02490}, 
}

@misc{shao2024deepseekmathpushinglimitsmathematical,
      title={DeepSeekMath: Pushing the Limits of Mathematical Reasoning in Open Language Models}, 
      author={Zhihong Shao and Peiyi Wang and Qihao Zhu and Runxin Xu and Junxiao Song and Xiao Bi and Haowei Zhang and Mingchuan Zhang and Y. K. Li and Y. Wu and Daya Guo},
      year={2024},
      eprint={2402.03300},
      archivePrefix={arXiv},
      primaryClass={cs.CL},
      url={https://arxiv.org/abs/2402.03300}, 
}

@article{weininger1988smiles,
  title={SMILES, a chemical language and information system. 1. Introduction to methodology and encoding rules},
  author={Weininger, David},
  journal={Journal of chemical information and computer sciences},
  volume={28},
  number={1},
  pages={31--36},
  year={1988},
  publisher={ACS Publications}
}

@article{heller2013inchi,
  title={InChI-the worldwide chemical structure identifier standard},
  author={Heller, Stephen and McNaught, Alan and Stein, Stephen and Tchekhovskoi, Dmitrii and Pletnev, Igor},
  journal={Journal of cheminformatics},
  volume={5},
  pages={1--9},
  year={2013},
  publisher={Springer}
}

@article{staker2019molecular,
  title={Molecular structure extraction from documents using deep learning},
  author={Staker, Joshua and Marshall, Kyle and Abel, Robert and McQuaw, Carolyn M},
  journal={Journal of chemical information and modeling},
  volume={59},
  number={3},
  pages={1017--1029},
  year={2019},
  publisher={ACS Publications}
}

@article{clevert2021img2mol,
  title={Img2Mol--accurate SMILES recognition from molecular graphical depictions},
  author={Clevert, Djork-Arn{\'e} and Le, Tuan and Winter, Robin and Montanari, Floriane},
  journal={Chemical science},
  volume={12},
  number={42},
  pages={14174--14181},
  year={2021},
  publisher={Royal Society of Chemistry}
}

@article{weir2021chempix,
  title={ChemPix: automated recognition of hand-drawn hydrocarbon structures using deep learning},
  author={Weir, Hayley and Thompson, Keiran and Woodward, Amelia and Choi, Benjamin and Braun, Augustin and Mart{\'\i}nez, Todd J},
  journal={Chemical science},
  volume={12},
  number={31},
  pages={10622--10633},
  year={2021},
  publisher={Royal Society of Chemistry}
}

@article{yi2022micer,
  title={MICER: a pre-trained encoder--decoder architecture for molecular image captioning},
  author={Yi, Jiacai and Wu, Chengkun and Zhang, Xiaochen and Xiao, Xinyi and Qiu, Yanlong and Zhao, Wentao and Hou, Tingjun and Cao, Dongsheng},
  journal={Bioinformatics},
  volume={38},
  number={19},
  pages={4562--4572},
  year={2022},
  publisher={Oxford University Press}
}

@article{rajan2021decimer,
  title={DECIMER 1.0: deep learning for chemical image recognition using transformers},
  author={Rajan, Kohulan and Zielesny, Achim and Steinbeck, Christoph},
  journal={Journal of Cheminformatics},
  volume={13},
  pages={1--16},
  year={2021},
  publisher={Springer}
}

@article{xu2022swinocsr,
  title={SwinOCSR: end-to-end optical chemical structure recognition using a Swin Transformer},
  author={Xu, Zhanpeng and Li, Jianhua and Yang, Zhaopeng and Li, Shiliang and Li, Honglin},
  journal={Journal of Cheminformatics},
  volume={14},
  number={1},
  pages={41},
  year={2022},
  publisher={Springer}
}

@article{campos2021img2smi,
  title={IMG2SMI: translating molecular structure images to simplified molecular-input line-entry system},
  author={Campos, Daniel and Ji, Heng},
  journal={arXiv preprint arXiv:2109.04202},
  year={2021}
}

@article{khokhlov2022image2smiles,
  title={Image2SMILES: Transformer-based molecular optical recognition engine},
  author={Khokhlov, Ivan and Krasnov, Lev and Fedorov, Maxim V and Sosnin, Sergey},
  journal={Chemistry-Methods},
  volume={2},
  number={1},
  pages={e202100069},
  year={2022},
  publisher={Wiley Online Library}
}

@article{li2024image2inchi,
  title={Image2InChI: Automated Molecular Optical Image Recognition},
  author={Li, Da-zhou and Xu, Xin and Pan, Jia-heng and Gao, Wei and Zhang, Shi-rui},
  journal={Journal of Chemical Information and Modeling},
  volume={64},
  number={9},
  pages={3640--3649},
  year={2024},
  publisher={ACS Publications}
}

@article{fang2024molparser,
  title={MolParser: end-to-end visual recognition of molecule structures in the wild},
  author={Fang, Xi and Wang, Jiankun and Cai, Xiaochen and Chen, Shangqian and Yang, Shuwen and Tao, Haoyi and Wang, Nan and Yao, Lin and Zhang, Linfeng and Ke, Guolin},
  journal={arXiv preprint arXiv:2411.11098},
  year={2024}
}

@misc{li2025chemvlmexploringpowermultimodal,
      title={ChemVLM: Exploring the Power of Multimodal Large Language Models in Chemistry Area}, 
      author={Junxian Li and Di Zhang and Xunzhi Wang and Zeying Hao and Jingdi Lei and Qian Tan and Cai Zhou and Wei Liu and Yaotian Yang and Xinrui Xiong and Weiyun Wang and Zhe Chen and Wenhai Wang and Wei Li and Shufei Zhang and Mao Su and Wanli Ouyang and Yuqiang Li and Dongzhan Zhou},
      year={2025},
      eprint={2408.07246},
      archivePrefix={arXiv},
      primaryClass={cs.LG},
      url={https://arxiv.org/abs/2408.07246}, 
}

@article{Zhao_2024,
   title={ChemDFM-X: towards large multimodal model for chemistry},
   volume={67},
   ISSN={1869-1919},
   url={http://dx.doi.org/10.1007/s11432-024-4243-0},
   DOI={10.1007/s11432-024-4243-0},
   number={12},
   journal={Science China Information Sciences},
   publisher={Springer Science and Business Media LLC},
   author={Zhao, Zihan and Chen, Bo and Li, Jingpiao and Chen, Lu and Wen, Liyang and Wang, Pengyu and Zhu, Zichen and Zhang, Danyang and Li, Yansi and Dai, Zhongyang and Chen, Xin and Yu, Kai},
   year={2024},
   month=dec }

@misc{fan2025ocsuopticalchemicalstructure,
      title={OCSU: Optical Chemical Structure Understanding for Molecule-centric Scientific Discovery}, 
      author={Siqi Fan and Yuguang Xie and Bowen Cai and Ailin Xie and Gaochao Liu and Mu Qiao and Jie Xing and Zaiqing Nie},
      year={2025},
      eprint={2501.15415},
      archivePrefix={arXiv},
      primaryClass={cs.CV},
      url={https://arxiv.org/abs/2501.15415}, 
}

@inproceedings{bukhari2019chemical,
  title={Chemical structure recognition (CSR) system: automatic analysis of 2D chemical structures in document images},
  author={Bukhari, Syed Saqib and Iftikhar, Zaryab and Dengel, Andreas},
  booktitle={2019 International Conference on Document Analysis and Recognition (ICDAR)},
  pages={1262--1267},
  year={2019},
  organization={IEEE}
}

@misc{filippov2009optical,
  title={Optical structure recognition software to recover chemical information: OSRA, an open source solution},
  author={Filippov, Igor V and Nicklaus, Marc C},
  year={2009},
  publisher={ACS Publications}
}

@article{mcdaniel1992kekule,
  title={Kekule: OCR-optical chemical (structure) recognition},
  author={McDaniel, Joe R and Balmuth, Jason R},
  journal={Journal of chemical information and computer sciences},
  volume={32},
  number={4},
  pages={373--378},
  year={1992},
  publisher={ACS Publications}
}

@inproceedings{ouyang2011chemink,
  title={Chemink: a natural real-time recognition system for chemical drawings},
  author={Ouyang, Tom Y and Davis, Randall},
  booktitle={Proceedings of the 16th international conference on Intelligent user interfaces},
  pages={267--276},
  year={2011}
}

@inproceedings{peryea2019molvec,
  title={MOLVEC: Open source library for chemical structure recognition},
  author={Peryea, Tyler and Katzel, Daniel and Zhao, Tongan and Southall, Noel and Nguyen, Dac-Trung},
  booktitle={Abstracts of papers of the American Chemical Society},
  volume={258},
  year={2019},
  organization={Amer Chemical Soc 1155 16TH ST, NW, WASHINGTON, DC 20036 USA}
}

@inproceedings{sadawi2012chemical,
  title={Chemical structure recognition: a rule-based approach},
  author={Sadawi, Noureddin M and Sexton, Alan P and Sorge, Volker},
  booktitle={Document recognition and retrieval XIX},
  volume={8297},
  pages={101--109},
  year={2012},
  organization={SPIE}
}

@inproceedings{smolov2011imago,
  title={Imago: Open-Source Toolkit for 2D Chemical Structure Image Recognition.},
  author={Smolov, Viktor and Zentsev, Fedor and Rybalkin, Mikhail},
  booktitle={TREC},
  year={2011}
}

@article{oldenhof2020chemgrapher,
  title={ChemGrapher: optical graph recognition of chemical compounds by deep learning},
  author={Oldenhof, Martijn and Arany, Adam and Moreau, Yves and Simm, Jaak},
  journal={Journal of chemical information and modeling},
  volume={60},
  number={10},
  pages={4506--4517},
  year={2020},
  publisher={ACS Publications}
}

@article{xu2022molminer,
  title={MolMiner: you only look once for chemical structure recognition},
  author={Xu, Youjun and Xiao, Jinchuan and Chou, Chia-Han and Zhang, Jianhang and Zhu, Jintao and Hu, Qiwan and Li, Hemin and Han, Ningsheng and Liu, Bingyu and Zhang, Shuaipeng and others},
  journal={Journal of Chemical Information and Modeling},
  volume={62},
  number={22},
  pages={5321--5328},
  year={2022},
  publisher={ACS Publications}
}

@article{zhang2022abc,
  title={ABC-Net: a divide-and-conquer based deep learning architecture for SMILES recognition from molecular images},
  author={Zhang, Xiao-Chen and Yi, Jia-Cai and Yang, Guo-Ping and Wu, Cheng-Kun and Hou, Ting-Jun and Cao, Dong-Sheng},
  journal={Briefings in bioinformatics},
  volume={23},
  number={2},
  pages={bbac033},
  year={2022},
  publisher={Oxford University Press}
}

@techreport{qian2022robust,
  title={Robust molecular image recognition: A graph generation approach},
  author={Qian, Yujie and Tu, Zhengkai and Guo, Jiang and Coley, Connor W and Barzilay, Regina},
  year={2022},
  institution={Technical Report}
}

@inproceedings{yoo2022image,
  title={Image-to-graph transformers for chemical structure recognition},
  author={Yoo, Sanghyun and Kwon, Ohyun and Lee, Hoshik},
  booktitle={ICASSP 2022-2022 IEEE International Conference on Acoustics, Speech and Signal Processing (ICASSP)},
  pages={3393--3397},
  year={2022},
  organization={IEEE}
}

@inproceedings{morin2023molgrapher,
  title={MolGrapher: graph-based visual recognition of chemical structures},
  author={Morin, Lucas and Danelljan, Martin and Agea, Maria Isabel and Nassar, Ahmed and Weber, Valery and Meijer, Ingmar and Staar, Peter and Yu, Fisher},
  booktitle={Proceedings of the IEEE/CVF International Conference on Computer Vision},
  pages={19552--19561},
  year={2023}
}

@article{brinkhaus2022randepict,
  title={RanDepict: Random chemical structure depiction generator},
  author={Brinkhaus, Henning Otto and Rajan, Kohulan and Zielesny, Achim and Steinbeck, Christoph},
  journal={Journal of cheminformatics},
  volume={14},
  number={1},
  pages={31},
  year={2022},
  publisher={Springer}
}

@article{brinkhaus2022decimer,
  title={DECIMER—hand-drawn molecule images dataset},
  author={Brinkhaus, Henning Otto and Zielesny, Achim and Steinbeck, Christoph and Rajan, Kohulan},
  journal={Journal of Cheminformatics},
  volume={14},
  number={1},
  pages={36},
  year={2022},
  publisher={Springer}
}

@article{rajan2020review,
  title={A review of optical chemical structure recognition tools},
  author={Rajan, Kohulan and Brinkhaus, Henning Otto and Zielesny, Achim and Steinbeck, Christoph},
  journal={Journal of Cheminformatics},
  volume={12},
  pages={1--13},
  year={2020},
  publisher={Springer}
}

@misc{staker2018molecularstructureextractiondocuments,
      title={Molecular Structure Extraction From Documents Using Deep Learning}, 
      author={Joshua Staker and Kyle Marshall and Robert Abel and Carolyn McQuaw},
      year={2018},
      eprint={1802.04903},
      archivePrefix={arXiv},
      primaryClass={cs.LG},
      url={https://arxiv.org/abs/1802.04903}, 
}

@article{morin2025markushgrapher,
  title={MarkushGrapher: Joint Visual and Textual Recognition of Markush Structures},
  author={Morin, Lucas and Weber, Val{\'e}ry and Nassar, Ahmed and Meijer, Gerhard Ingmar and Van Gool, Luc and Li, Yawei and Staar, Peter},
  journal={arXiv preprint arXiv:2503.16096},
  year={2025}
}

@article{qwen25vl,
  title={Qwen2. 5-vl technical report},
  author={Bai, Shuai and Chen, Keqin and Liu, Xuejing and Wang, Jialin and Ge, Wenbin and Song, Sibo and Dang, Kai and Wang, Peng and Wang, Shijie and Tang, Jun and others},
  journal={arXiv preprint arXiv:2502.13923},
  year={2025}
}

@article{Rajan2020,
author = {Rajan, Kohulan and Brinkhaus, Henning Otto and Zielesny, Achim and Steinbeck, Christoph},
doi = {10.1186/s13321-020-00465-0},
issn = {1758-2946},
journal = {Journal of Cheminformatics},
pages = {1--13},
publisher = {Springer International Publishing},
title = {{A review of optical chemical structure recognition tools}},
url = {https://doi.org/10.1186/s13321-020-00465-0},
year = {2020}
}

@misc{openai_gpt4o-mini,
  author = {OpenAI},
  title = {GPT-4o mini: advancing cost-efficient intelligence},
  howpublished = {\url{https://openai.com/index/gpt-4o-mini-advancing-cost-efficient-intelligence/}},
  year={2024}
}

\clearpage
\newpage
\beginappendix

\section{Dataset}
\label{app_dataset}

Building on these insights mentioned in Section 3 of the main paper, we developed \textbf{GTR-1.3M}, a specialized SFT dataset for VLM-based OCSR tasks. 
Following MolScribe and MolNexTR, \textbf{GTR-1.3M} is composed of two parts:
(1) \textbf{GTR-PubChem-1M}: We selected 1 million molecular SMILES from the PubChem database and used the Indigo tool to convert them into molecular images. 
The difference is that we chose the offline generation method to save the generated data locally as a way to accelerate the training process.
We precisely recorded the spatial positions of each atom and bond to construct the Graph Traversal process as the Chain of Thought (CoT). 
Following~\cite{qian2023molscribe}, we replaced specific functional groups with abbreviations to create superatoms and randomly added common R-group labels (R, R1, R2, R', etc.) to the molecules.
(2) \textbf{GTR-USPTO-351K}: This subset was created from USPTO-680K. 
We developed a data correction pipeline to correct and filter abbreviated structures in these samples, obtained 351k high-quality samples, and formed the \textbf{GTR-USPTO-351K} subset.

\subsection{Graph Traversal as Visual CoT}

\begin{figure}[h]
    \centering
    {\setlength{\fboxrule}{1.0pt}\color{white}
    \fbox{\includegraphics[width=0.5\linewidth]{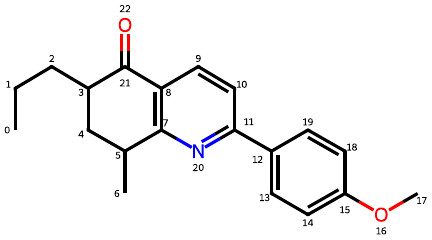}}}
    \caption{An example of our graph traversal order. The numbers indicate the traversal order of atoms.}
    \label{fig:graph_traversal}
\end{figure}

As shown in Figure~\ref{fig:graph_traversal}, the traversal starts with carbon atom 0, followed by carbon atom 1. Subsequently, the bond between atoms 0 and 1 is traversed. The process continues with carbon atom 2 and the bond between atoms 1 and 2. Following this pattern, the entire molecular graph is traversed step by step. This depth-first traversal strategy tends to prioritize branches with shallower depths.

\subsection{Construction of GTR-USPTO-351K}

\begin{figure*}[h]
    \centering
    \includegraphics[width=1\linewidth]{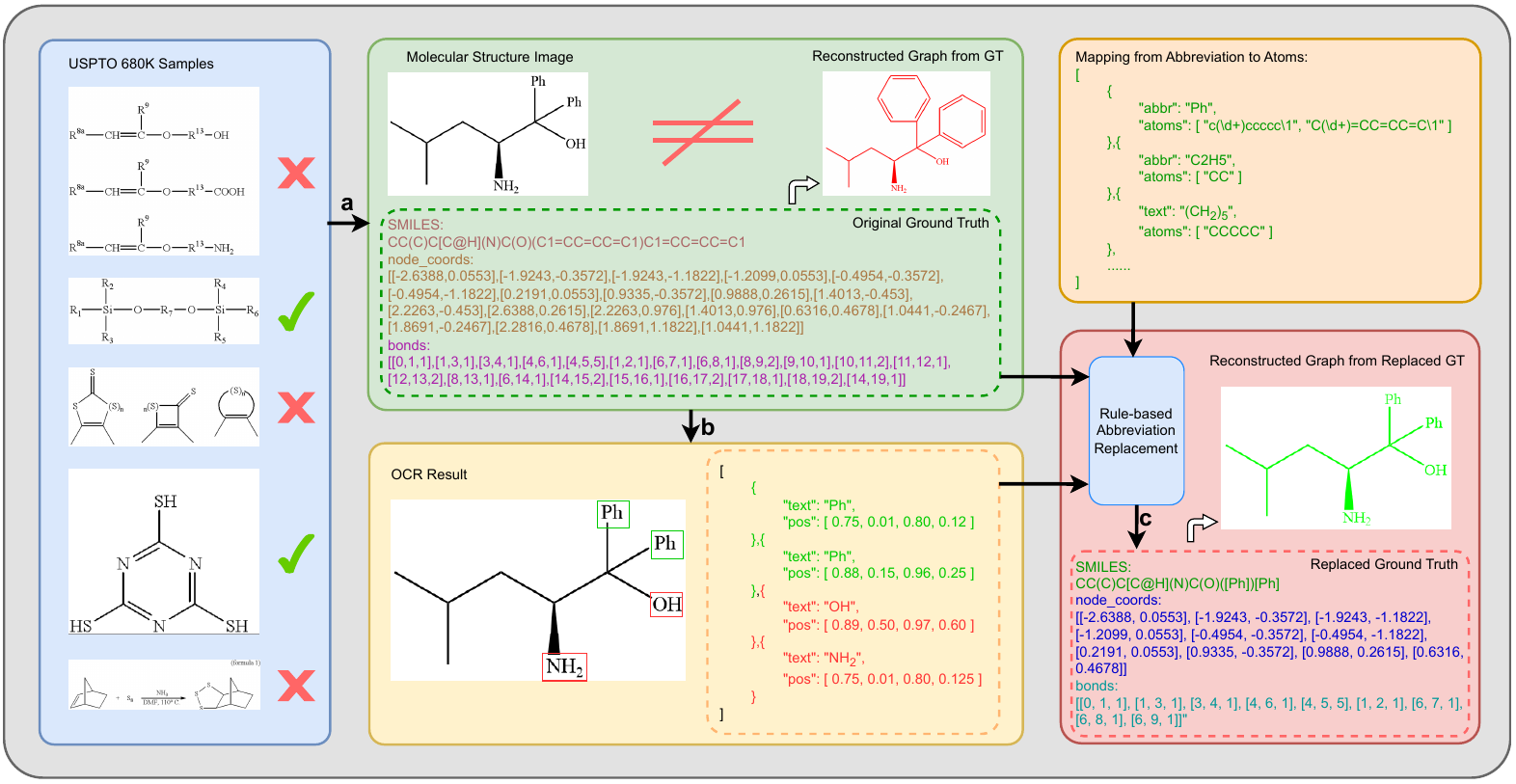}
    \caption{Due to the discrepancies between molecular structure images and their corresponding ground truth (SMILES\cite{weininger1988smiles}, node\_coords, and bonds), it is necessary to correct the ground truth in the USPTO-680K\cite{Rajan2020} dataset. The process is illustrated in Figure \ref{fig:dataset_construction}. In step \textbf{a}, a subset of 365k samples, each containing a single molecule, is selected from the USPTO-680K dataset. In step \textbf{b}, Optical Character Recognition (OCR) is applied to identify all characters in the molecular structure images. In step \textbf{c}, abbreviations identified by OCR that appear in the abbreviation-superatom mapping table are flagged for replacement in the ground truth. The replacement algorithm is rule-based, and its core logic determines whether the atomic combinations corresponding to each abbreviation are present in the SMILES notation, thereby guiding the decision to replace them.
The input required for the replacement algorithm includes the original SMILES, node\_coords, bonds, the OCR results, and the abbreviation-atom mapping table, which contains common functional groups in organic chemistry. The algorithm then applies the necessary substitutions to the SMILES, node\_coords, and bonds to ensure consistency between the molecular structure and the ground truth. Of the 365k samples, 351k had their ground truth successfully corrected.}
    \label{fig:dataset_construction}
\end{figure*}

As shown in the Figure \ref{fig:dataset_construction}, the rectification and filtering pipeline for USPTO datasets involves:
\textbf{(1) Image Screening:} Original images from USPTO-680K are preprocessed to remove samples with multiple molecular structures, ensuring each image corresponds to a single structure to avoid ambiguity.
\textbf{(2) Abbreviation Extraction:} OCR technology identifies and extracts chemical abbreviations (e.g., "Ph", "Et") from images for structural alignment. 
\textbf{(3) Structure Replacement and Mapping:} Using the extracted abbreviations, SMILES strings, atom sequences, and bond information in the ground truth are replaced. A mapping table between common abbreviations and their atomic structures guides this replacement. 
This process ensures high semantic consistency between image content and structural annotations, providing a reliable foundation for model learning.
After this process, we obtained 351k high-quality samples, forming the \textbf{GTR-USPTO-351K} subset.

\subsection{Construction of GTR-PubChem-1M}

Following MolScribe\cite{qian2023molscribe} and MolNexTR\cite{chen2024molnextr}, we saved the images generated during their training process, because the various data enhancement operations in them may cause RDKit~\footnote{http://www.rdkit.org/} to fail to render the correct images based on the generated SMILES. Therefore, although our total number of seed SMILES is 1,000,000, the final total number of samples obtained is 999,950. Details of the data enhancement and generation process can be found in MolScribe.

\subsection{Comparison of GTR-USPTO-351K and USPTO-680K}

\begin{figure*}[h]
    \centering
    \includegraphics[width=1\linewidth]{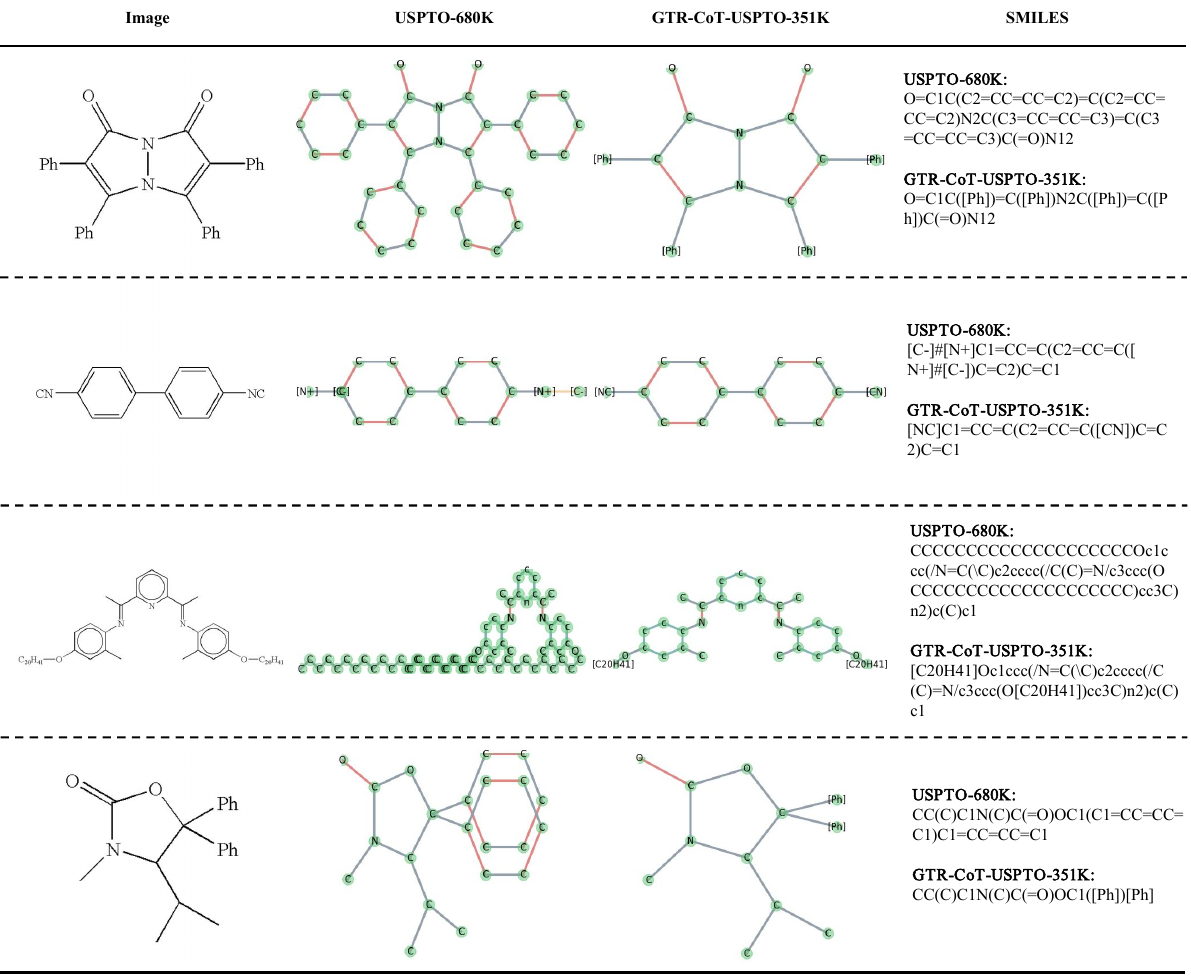}
    \caption{The visualization result of USPTO-680K and \textbf{GTR-USPTO-351K}. The reconstructed graphs of USPTO-351K (the second column) are unaligned with the molecular structure images. \textbf{GTR-USPOTO-351K}, however, strictly follow the molecular structure in the images. We use different colors to mark SINGLE, DOUBLE, TRIPLE, and AROMATIC bonds, as well as different colored arrows for BEGINWEDGE and BEGINDASH bonds.}
    \label{fig:dataset_vis}
\end{figure*}

Figure \ref{fig:dataset_vis} compares the ground truth of \textbf{GTR-USPTO-351K} and USPTO-680K. The strict alignment between molecular structure images and reconstructed graphs entails precise prediction results. 

\section{Evaluation}
\label{app_evaluation}

\subsection{Evaluation details of OCSR methods}

\begin{figure*}[h]
    \centering
    \includegraphics[width=1\linewidth]{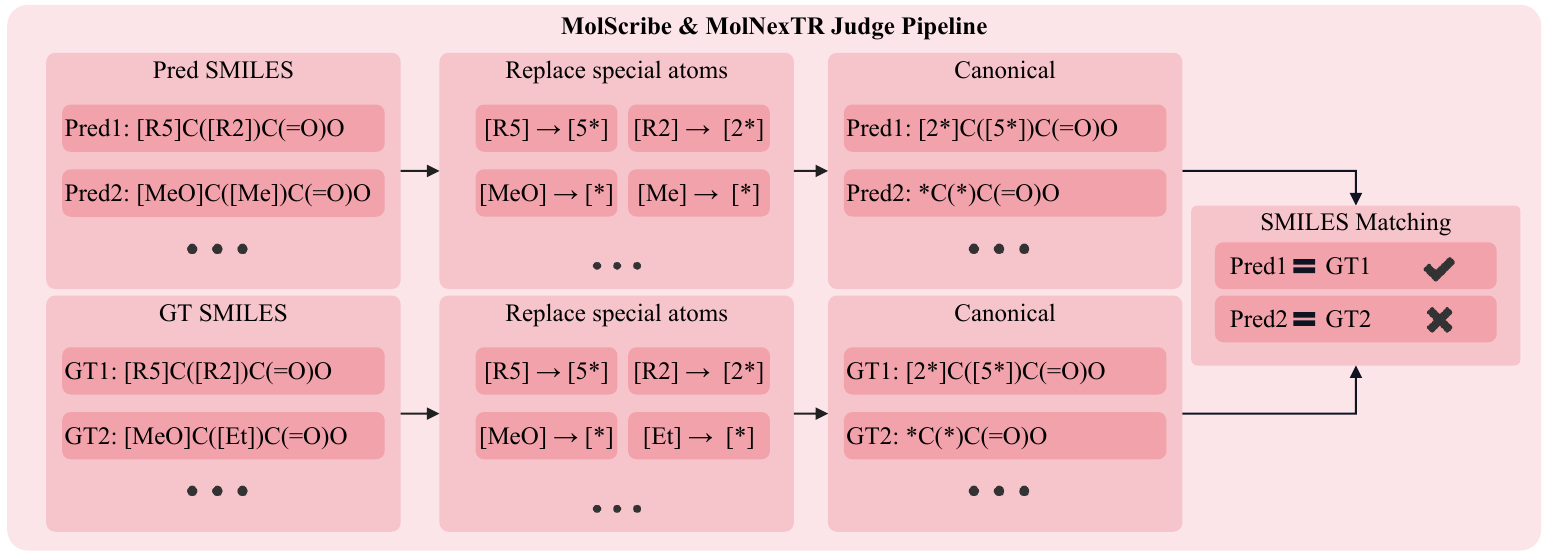}
    \caption{The evaluation process of MolScribe and MolNexTR. All the abbreviations of functional groups are replaced by * when comparing the predicted and ground truth SMILES.}
    \label{fig:molscribe_pipeline}
\end{figure*}

Figure \ref{fig:molscribe_pipeline} demonstrates the details of the evaluation of Molscribe and MolNexTR. The abbreviations of the functional groups occurring in the molecular structure images are simply replaced by * during the evaluation, resulting in the degradation of the evaluation accuracy.

\subsection{Graph-based Metric}

\begin{figure*}[h]
    \centering
    \includegraphics[width=1\linewidth]{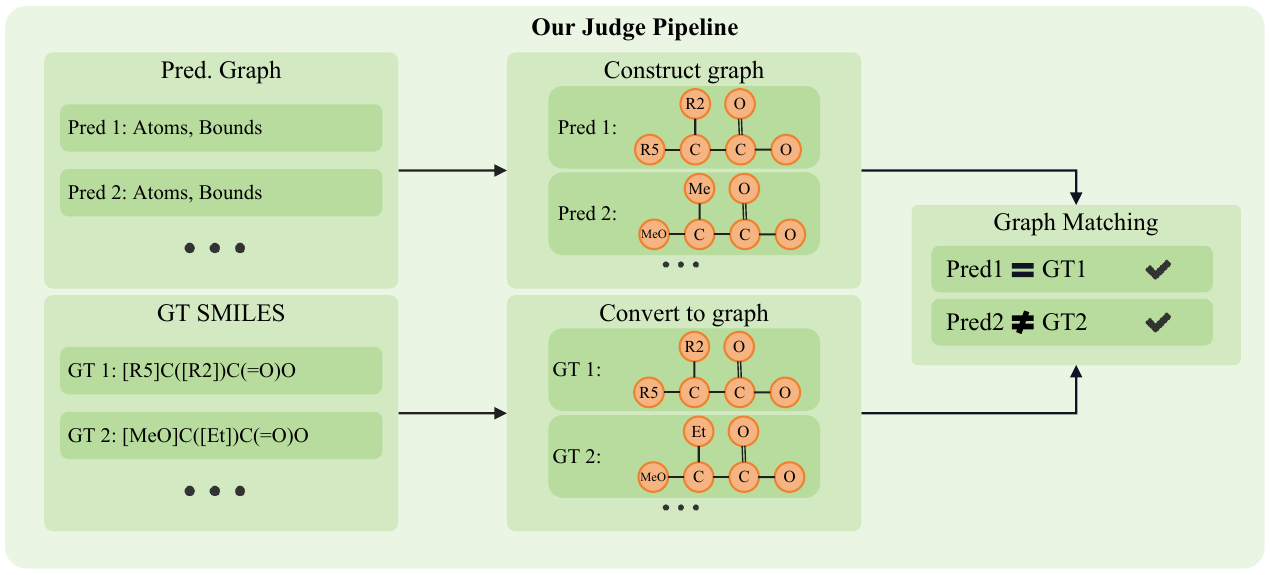}
    \caption{The evaluation process of our Graph-based metric. All the functional groups and Markush structures are kept in their abbreviations when comparing the predicted and ground truth graphs.}
    \label{fig:our_pipeline}
\end{figure*}

Figure \ref{fig:our_pipeline} demonstrates the details of the evaluation of our graph-based metric. The abbreviations of the functional groups are kept as it is. Meanwhile, the predicted and ground truth graphs are compared directly, instead of comparing the SMILES generated by these graphs, leading to a more accurate and direct evaluation diagram.

\section{More Experiment Results}
\label{app_more_exper_result}

\subsection{More Prediction Visualization}

\begin{figure*}[h]
    \centering
    \includegraphics[width=1.0\linewidth]{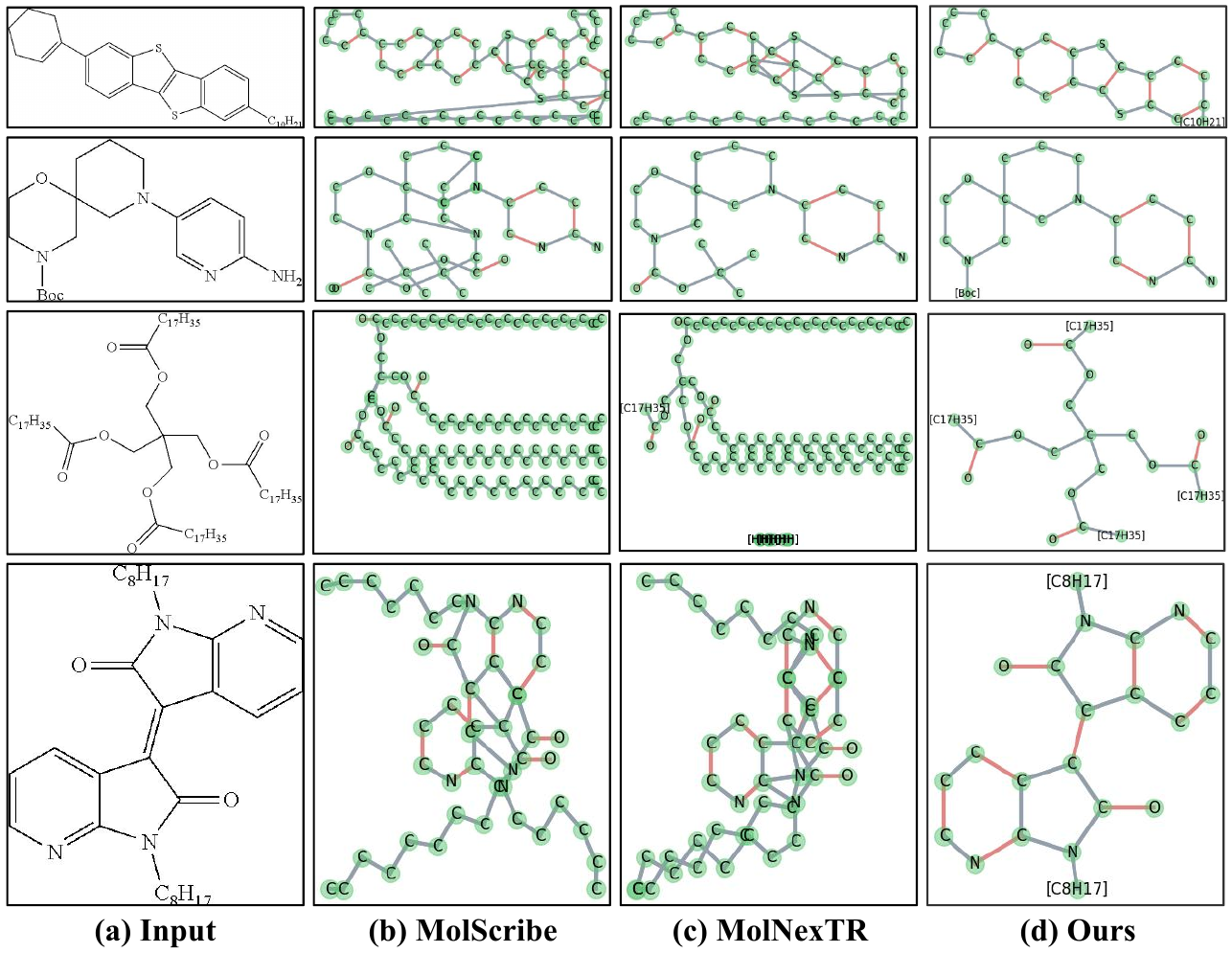}
    \caption{More prediction visualization of MolScribe, MolNexTR, and our method. We use different colors to mark SINGLE bonds, DOUBLE bonds, TRIPLE bonds, and AROMATIC bonds, as well as different colored arrows for BEGINWEDGE bonds and BEGINDASH bonds.}
    \label{fig:pred_vis}
\end{figure*}

Figure \ref{fig:pred_vis} shows more prediction visualization results of MolScribe, MolNexTR, and our method.

\subsection{Bad Case Analysis}

\begin{figure*}[h]
    \centering
    \includegraphics[width=1.0\linewidth]{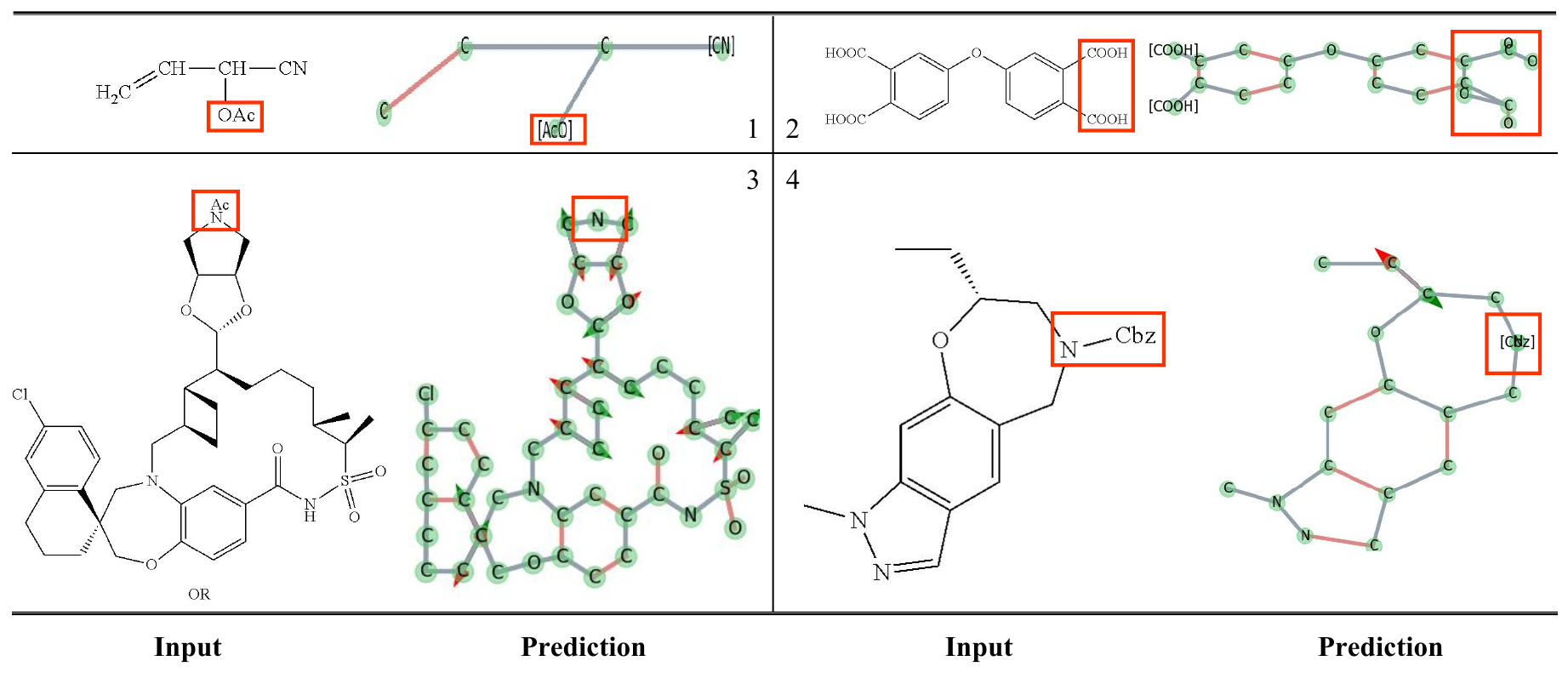}
    \caption{Some typical bad cases of our method on \textbf{MolRec-Abb}. We use different colors to mark SINGLE bonds, DOUBLE bonds, TRIPLE bonds, and AROMATIC bonds, as well as different colored arrows for BEGINWEDGE bonds and BEGINDASH bonds.}
    \label{fig:bad_case}
\end{figure*}

Figure \ref{fig:bad_case} shows some failure cases of our method on \textbf{MolRec-Abb}. In case 1, there is a problem of incorrectly predicting the order of the abbreviations. In case 2. The left side predicted "HOOC" retains the abbreviated form, but on the right side, "COOH" does the expansion form. In addition. In case 3, the model misses the topmost "Ac". This is mainly because there are too few samples of such a writing style, and the model has difficulty in predicting the abbreviations above and below most of the time. In case 4, the superatom "Cbz" on the far right is repeatedly positioned on a carbon atom.

\begin{figure*}[h]
    \centering
    \includegraphics[width=0.9\linewidth]{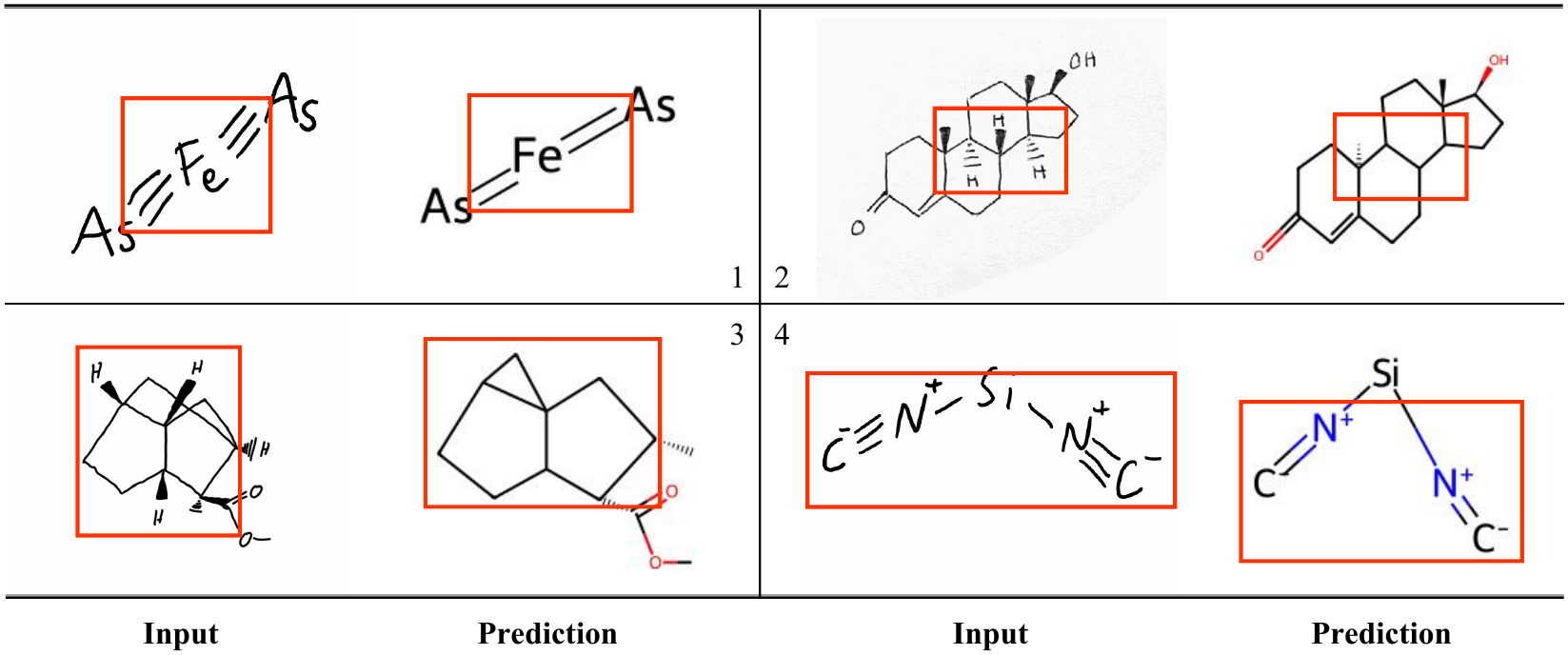}
    \caption{Some typical bad cases of our method on \textbf{DECIMER-HD-Test}. We use different colors to mark SINGLE bonds, DOUBLE bonds, TRIPLE bonds, and AROMATIC bonds, as well as different colored arrows for BEGINWEDGE bonds and BEGINDASH bonds.}
    \label{fig:bad_case_hd}
\end{figure*}

Figure \ref{fig:bad_case_hd} shows some failure cases of our method in \textbf{DECIMER-HD-Test}. 
In Case 1 and Case 2, our three-dimensional model failed to correctly predict the chiral hydrogen bond, which may be due to the small number of triple bond samples in the training dataset.
In Case 2 and Case 3, the model failed to predict the chiral hydrogen bond, which is the main direction of our future research.

\section{Chemoinformatics Basics}
\label{app_chemoinfo}

\subsection{SMILES}

SMILES is a method for representing the structure of a molecule as an ASCII string. David Weininger initially proposed the concept in the 1980s to represent and store information about chemical molecules in computers. The fundamental principle of SMILES is to describe the molecular topology (i.e., the manner in which atoms are interconnected) in a single line of string. This representation has applications in areas such as databases, machine learning, molecular searching, and cheminformatics.

Common organic atoms can be used directly with their atomic symbols (e.g. "C", "N", "O", "S", "P", "F", "Cl", "Br", "I"). Special or charged atoms are denoted using square brackets, for example, "[Na+]", "[Fe+3]", and "[C@H]". 
In SMILES, single bonds are typically not represented, double bonds are indicated by "=", triple bonds by "\#", and aromatic bonds by ":". Such as ethane: "CC", ethylene: "C=C", and acetylene: "C\#C". In SMILES, parentheses are frequently employed to denote branched chains, such as isopropanol: "CC(C)O".
Numbers are employed in SMILES to denote the commencement and cessation of atoms, thus forming a ring. For example: cyclohexane: "C1CCCC1", benzene: "c1ccccc1", (the lowercase "c" represents aromatic carbon). 
SMILES also supports the description of chiral centres; for example, "[C@H]" denotes a clockwise configuration, while "[C@@H]" denotes a counterclockwise configuration.

In comparison with structural formulae or molecular diagrams, SMILES representations are characterised by their conciseness and ease of use for database storage, searching, and chemical calculations. Furthermore, the majority of SMILES can be reduced to structural formulas and are supported by numerous chemical software applications (e.g., RDKit, Open Babel\footnote{http://openbabel.org/}).

\subsection{SMILES Canonicalization}
SMILES canonicalization is defined as the process of converting a molecular structure into a unique and canonical SMILES representation. "CCO" and  "OCC" are correct for ethanol, but they are frequently required to possess a single, distinct SMILES for a given molecule within databases or algorithms. In this instance, the utilisation of canonical SMILES is imperative to establish a benchmark. The implementation of canonical SMILES is essential for the standardisation of SMILES. This approach eliminates the need for duplicated data, expedites the process of retrieving and comparing information, and facilitates the learning and chemical modelling processes.
Implementing the canonicalization of SMILES is determined by the algorithm, not by the habits of human-written SMILES. It consists of 3 main steps:

\begin{enumerate}
    \item Molecular graph renumbering: View molecules as graph structures (atoms are nodes and bonds are edges) and renumber atoms, giving each atom an index.
    \item Topology-based sorting: sort the molecules according to their structural rules and chemical properties (e.g. atomic number, connectivity, ring structure, etc.) and choose a minimum or optimal order.
    \item Follow the path after sorting to generate a SMILES string in standard form.
\end{enumerate}

However, due to the uncertainty of the chemical nature of the abbreviated structures and the fact that they cannot be exhaustively enumerated to give a fixed index, it is not possible to distinguish between the two abbreviated structures when they are both located at the endpoints of the molecule. This results in an Incorrect judgment when using canonical SMILE for Exact Match.

\begin{figure*}[h]
    \centering
    \includegraphics[width=1\linewidth]{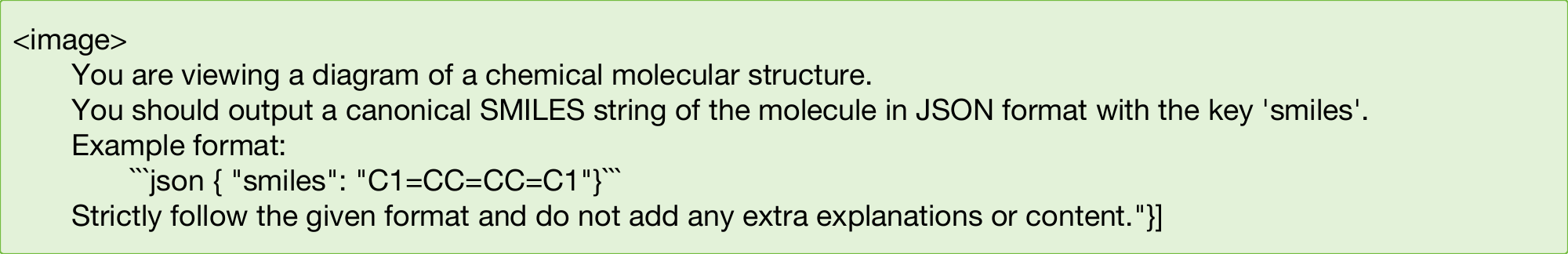}
    \caption{Prompt for our model to directly predict SMILES.}
    \label{fig:prompt_smiles}
\end{figure*}

\section{All prompts}
\label{app_all_prompts}

\subsection{GTR's Prompts}

\begin{figure*}[h]
    \centering
    \includegraphics[width=1\linewidth]{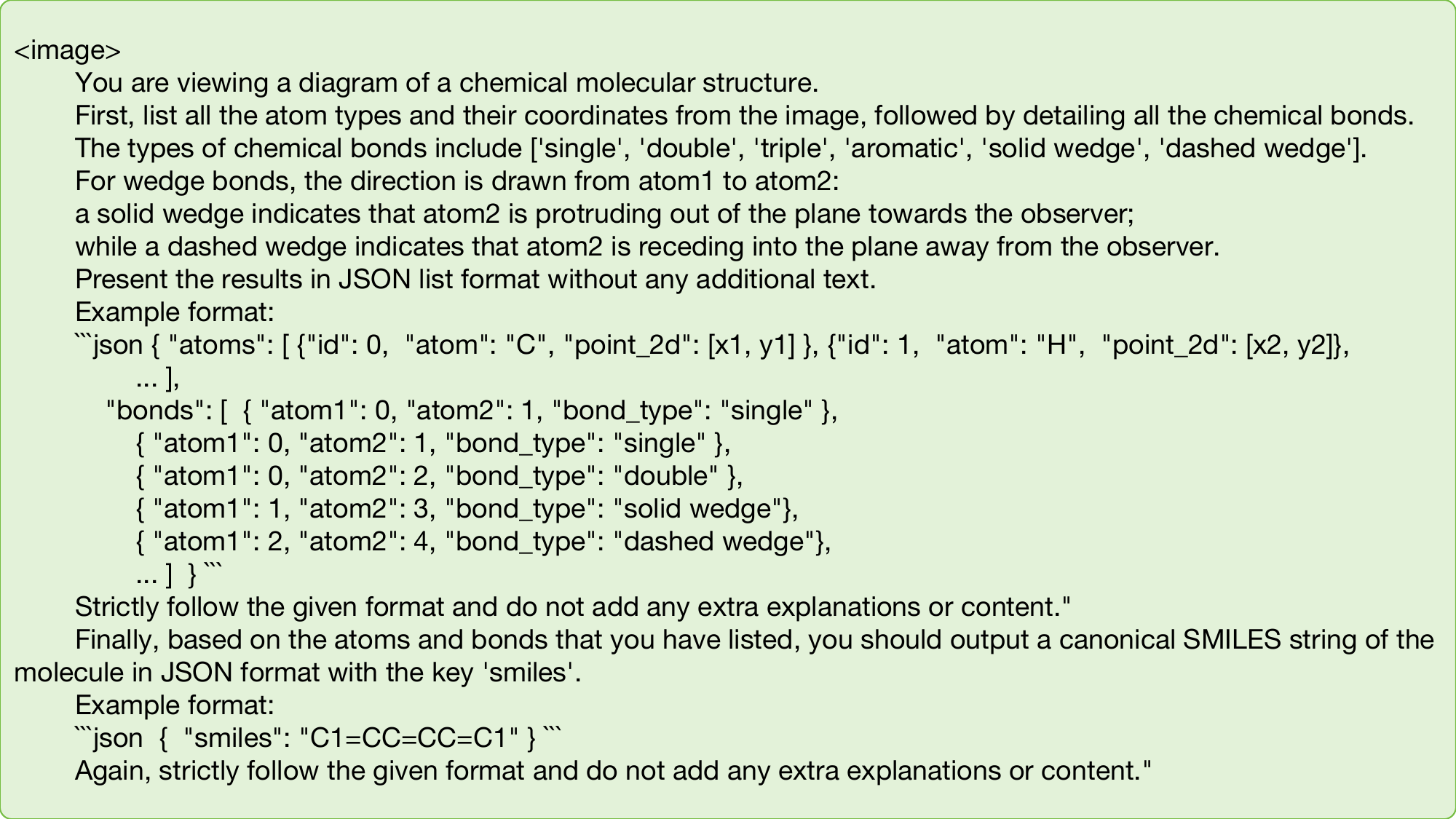}
    \caption{Prompt for our model to predict the atom first, then bonds, and finally SMILES.}
    \label{fig:prompt_v2}
\end{figure*}

\begin{figure*}[h]
    \centering
    \includegraphics[width=1\linewidth]{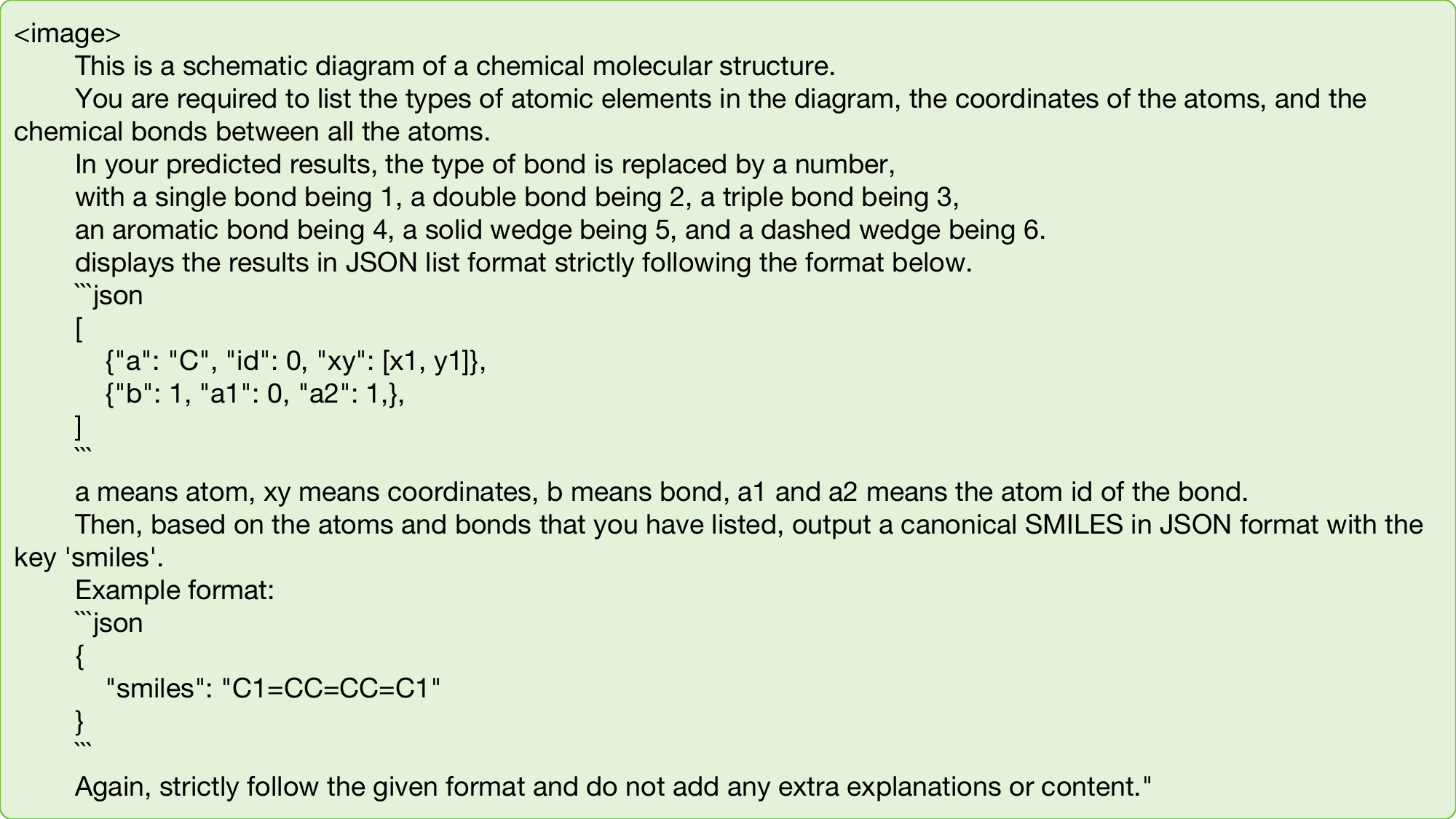}
    \caption{Prompt for our model to predict atoms and bonds in a graph traversal manner first, then predict SMILES (Our method).}
    \label{fig:prompt_v3}
\end{figure*}

The following three figures demonstrate the prompt used by our model.
Figure \ref{fig:prompt_smiles} demonstrates the prompt for the direct prediction.
Figure \ref{fig:prompt_v2} demonstrates the prompt for predicting the atoms first, then bonds, and finally SMILES.
Figure \ref{fig:prompt_v3} demonstrates the prompt for predicting the atoms and bonds in a graph traversal manner first, then predicting SMILES.

\subsection{Proprietary VLMs' Prompts}

\begin{figure*}[h]
    \centering
    \includegraphics[width=1\linewidth]{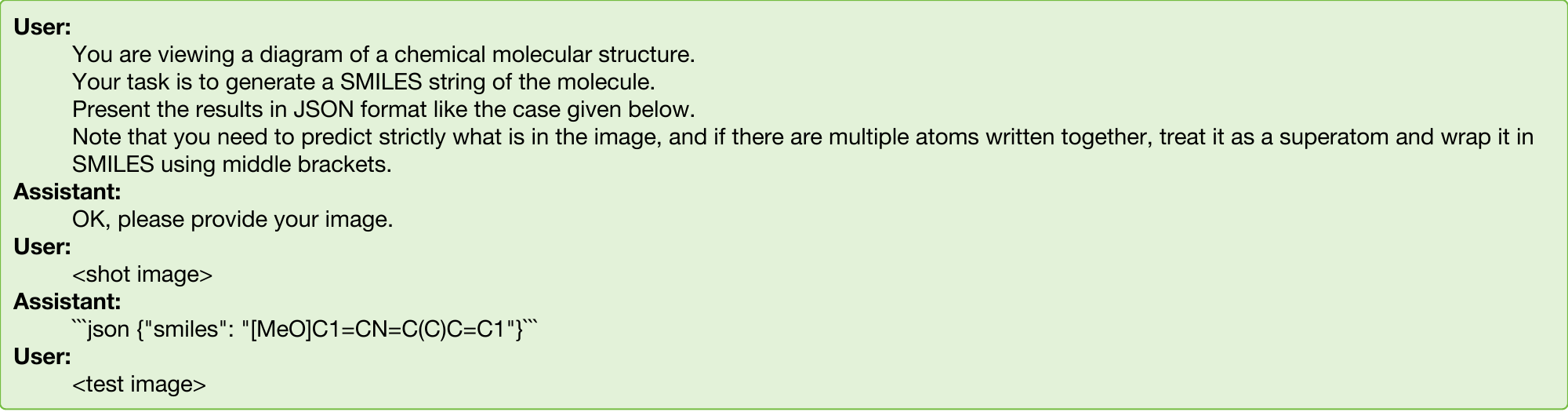}
    \caption{Prompt for proprietary LLMs to directly predict SMILES.}
    \label{fig:api_prompt_smiles}
\end{figure*}

\begin{figure*}[h]
    \centering
    \includegraphics[width=1\linewidth]{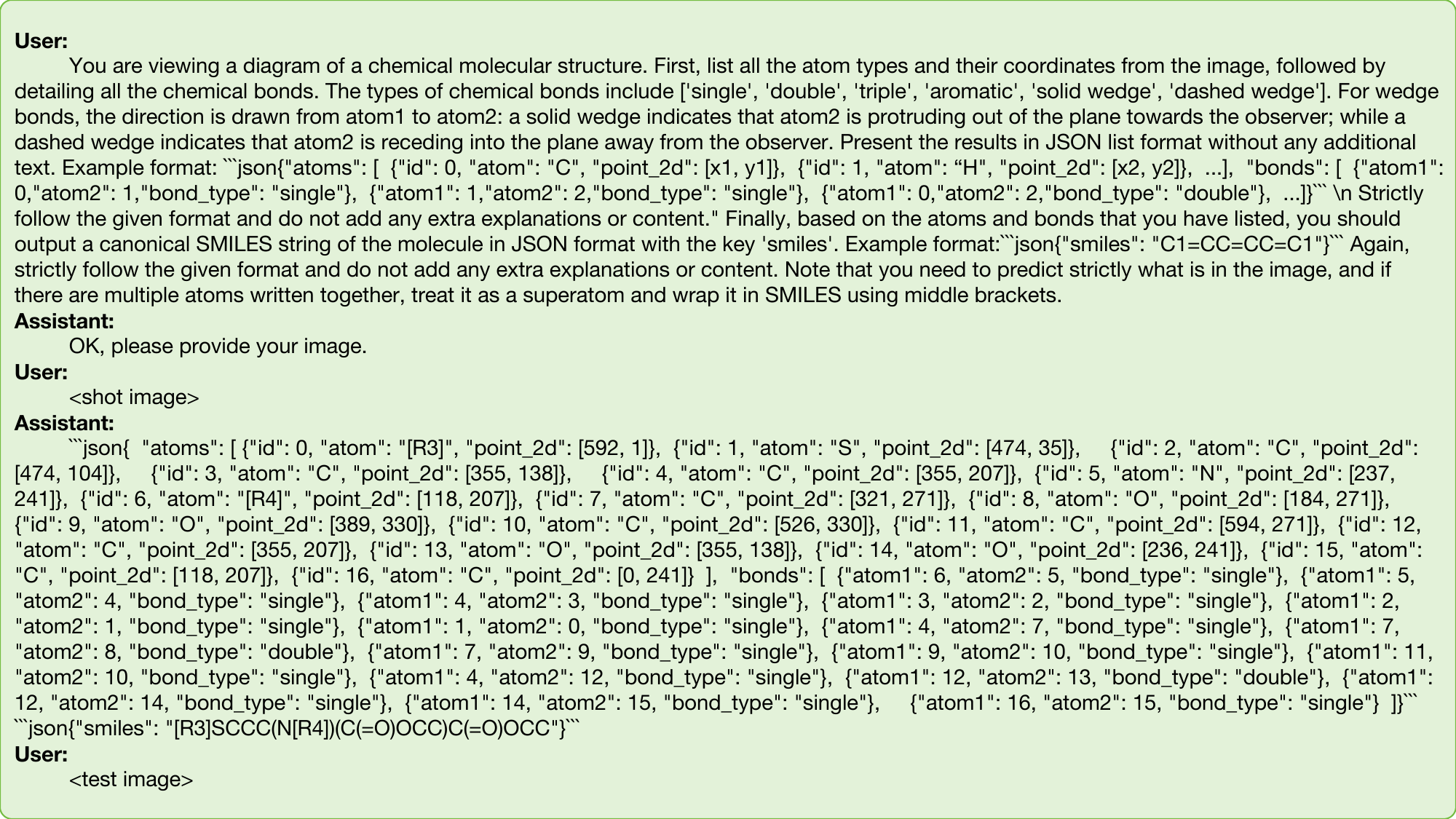}
    \caption{Prompt for proprietary LLMs to predict the atoms and bonds in a graph traversal manner first, then predicting SMILES.}
    \label{fig:api_prompt_v3}
\end{figure*}

The following two figures demonstrate the prompt used by three proprietary VLMs (Qwen-VL-Max-2025-04-08\cite{qwen25vl}, GPT-4o-mini-2024-07-18\cite{openai_gpt4o-mini}, and GPT-4o-2024-08-06\cite{hello_gpt-4o_2024}).
Figure \ref{fig:api_prompt_smiles} demonstrates the prompt for the direct prediction of SMILES.
Figure \ref{fig:api_prompt_v3} demonstrates the prompt for predicting the atoms and bonds in a graph traversal manner first, then predicting SMILES.

\section{Related Works}

\subsection{OCSR Methods}
\label{sec:ocsr_methods}

OCSR methods are generally categorized into image-captioning and graph-parsing approaches. Image-captioning methods treat OCSR as an image captioning task, directly outputting SMILES strings. These models use an encoder to extract visual features and a decoder to generate SMILES\cite{weininger1988smiles} or InChI\cite{heller2013inchi} sequences. Early models combined convolutional encoders with recurrent decoders (RNN, GRU, or LSTM), such as MSE-DUDL\cite{staker2019molecular}, DECIMER\cite{rajan2020decimer}, Img2Mol\cite{clevert2021img2mol}, ChemPix\cite{weir2021chempix}, and MICER\cite{yi2022micer}. Later works introduced transformer-based architectures, including DECIMER 1.0\cite{rajan2021decimer}, DECIMER 2.0\cite{rajan2023decimer}, SwinOCSR\cite{xu2022swinocsr}, IMG2SMI\cite{campos2021img2smi}, Image2SMILES\cite{khokhlov2022image2smiles}, Image2InChI\cite{li2024image2inchi}, and MolParser\cite{fang2024molparser}. Recently, vision-language models (VLM) have been applied to this task, as seen in ChemVLM\cite{li2025chemvlmexploringpowermultimodal}, ChemDFM-X\cite{Zhao_2024}, and OCSU\cite{fan2025ocsuopticalchemicalstructure}.

Graph-parsing methods predict atoms, bonds, and additional information (e.g., charges) from images to derive molecular graph structures. Early methods used hand-crafted rules for component detection and graph reconstruction\cite{bukhari2019chemical, filippov2009optical, mcdaniel1992kekule, ouyang2011chemink, peryea2019molvec, sadawi2012chemical, smolov2011imago}. While effective in simple, noise-free scenarios, these methods struggle with complexity and have high maintenance costs. Recent methods leverage deep learning for component detection or segmentation\cite{oldenhof2020chemgrapher, xu2022molminer, zhang2022abc}, and more recent approaches use deep learning to construct graphs directly\cite{qian2022robust, yoo2022image}.
Existing graph-parsing methods, such as MolGrapher\cite{morin2023molgrapher}, MolScribe\cite{qian2023molscribe}, and MolNexTr\cite{chen2024molnextr}, use a two-stage approach (Figure 1 of the main paper). They first predict atoms (nodes) and then chemical bonds (edges) using classifiers or graph neural networks. This approach results in redundant computations and increased task complexity.

\subsection{OCSR Datasets and Evaluation Benchmarks}
\label{sec:oscr_dataset_and_benchmark}
OCSR datasets and benchmarks are categorized into synthetic and real images. 
In synthetic datasets, tools like RDKit or Indigo generate molecular images from chemical database SMILES \cite{qian2023molscribe,chen2024molnextr,fang2024molparser}.
In \cite{rajan2024advancements}, RanDepict\cite{brinkhaus2022randepict} is used to synthesize images that simulate handwritten forms, enhancing model performance on actual handwritten test sets.

Real image datasets contain images from patents and literatures. For example, real data from USPTO patent documents are adopted in \cite{qian2023molscribe,chen2024molnextr}. \cite{brinkhaus2022decimer} is a real handwritten dataset containing 5088 samples.
Many evaluation datasets come from real patents or papers. The review article\cite{rajan2020review} organizes multiple evaluation sets from real patents or papers. \cite{staker2018molecularstructureextractiondocuments,qian2023molscribe,morin2023molgrapher} also introduce evaluation sets from USPTO patents or journals.
Most training sets and all evaluation sets focus on image-captioning methods, containing only images and corresponding SMILES. Thus, even graph-parsing methods rely on comparing predicted SMILES with ground truth, limiting direct evaluation.



\section{Limitation}

While \textbf{GTR-VLM} achieves strong results on MolRec-Abb and MolRec-USPTO, further improvements are possible through the integration of advanced reinforcement learning strategies and the development of higher-quality datasets. Despite these opportunities for enhancement, our approach constitutes a pioneering contribution to the field, offering a novel, robust solution for OCSR and providing meaningful guidance for downstream applications such as automated chemical literature analysis.

\end{document}